\documentclass[10pt,twocolumn,letterpaper]{article}

\usepackage{iccv}
\usepackage{times}
\usepackage{epsfig}
\usepackage{graphicx}
\usepackage{amsmath}
\usepackage{amssymb}
\usepackage{booktabs} 
\usepackage{bbold}
\usepackage{pifont}
\usepackage[toc,page,header]{appendix}
\usepackage{etoc}
\usepackage[accsupp]{axessibility} 

\usepackage[pagebackref=true,breaklinks=true,letterpaper=true,colorlinks,bookmarks=false]{hyperref}
 
\newcommand{\name}[1]{ViLLA}

\newcommand{\stoptocwriting}{\addtocontents{toc}{\protect\setcounter{tocdepth}{-5}}}
\newcommand{\resumetocwriting}{\addtocontents{toc}{\protect\setcounter{tocdepth}{\arabic{tocdepth}}}}

\newcommand{\cmark}{\ding{51}}%
\newcommand{\xmark}{\ding{55}}%

\iccvfinalcopy 

\def\iccvPaperID{10940}

\begin{document}

\title{\name{}: Fine-Grained Vision-Language Representation Learning from Real-World Data}

\author{Maya Varma \hspace{4mm} Jean-Benoit Delbrouck \hspace{4mm} Sarah Hooper \hspace{4mm} Akshay Chaudhari \hspace{4mm} Curtis Langlotz
\\
Stanford University\\
{\tt\small \{mvarma2, jbdel, smhooper, akshaysc, langlotz\}@stanford.edu}\\
}

\maketitle

\begin{abstract}
Vision-language models (VLMs), such as CLIP and ALIGN, are generally trained on datasets consisting of image-caption pairs obtained from the web. However, real-world multimodal datasets, such as healthcare data, are significantly more complex: each image (\eg X-ray) is often paired with text (\eg physician report) that describes many distinct attributes occurring in fine-grained regions of the image. We refer to these samples as exhibiting high pairwise complexity, since each image-text pair can be decomposed into a large number of region-attribute pairings. The extent to which VLMs can capture fine-grained relationships between image regions and textual attributes when trained on such data has not been previously evaluated. The first key contribution of this work is to demonstrate through systematic evaluations that as the pairwise complexity of the training dataset increases, standard VLMs struggle to learn region-attribute relationships, exhibiting performance degradations of up to 37\% on retrieval tasks. In order to address this issue, we introduce \name{} as our second key contribution. \name{}, which is trained to capture fine-grained region-attribute relationships from complex datasets, involves two components: (a) a lightweight, self-supervised mapping model to decompose image-text samples into region-attribute pairs, and (b) a contrastive VLM to learn representations from generated region-attribute pairs. We demonstrate with experiments across four domains (synthetic, product, medical, and natural images) that \name{} outperforms comparable VLMs on fine-grained reasoning tasks, such as zero-shot object detection (up to 3.6 AP50 points on COCO and 0.6 mAP points on LVIS) and retrieval (up to 14.2 R-Precision points)\footnote{Code: \url{https://github.com/StanfordMIMI/villa}}.

\end{abstract}

\stoptocwriting
\section{Introduction}
\label{sec:intro}

\begin{figure*}
\label{fig:header}
\begin{center}
\includegraphics[width=\textwidth,height=5.2cm]{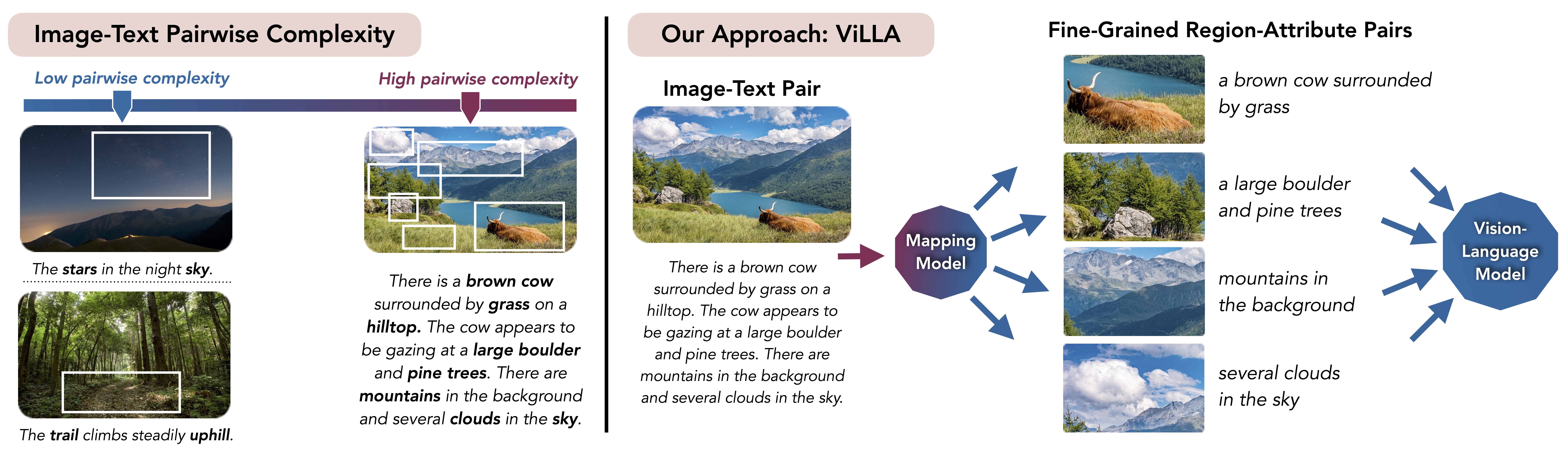}
\end{center}
  \caption{(Left) We provide examples of image-text samples with varying complexities. Examples with low pairwise complexity are from the CC3M dataset \cite{sharma-2018-conceptualcap}. Textual attributes are in bold type, and corresponding image regions are marked with bounding boxes. (Right) We introduce \name{}, a representation learning approach that captures fine-grained relationships between image regions and textual attributes.}
\label{fig:short}
\end{figure*}

Vision-language models (VLMs), which jointly learn relationships between images and text, have been shown in recent years to be highly effective on a variety of classification, retrieval, and robustness tasks \cite{radford-clip, jia-align,pham2021basic,domino2022,zhang2020convirt,cyclip2022}. VLMs are trained on large-scale datasets consisting of image-text pairs, where the text takes the form of a concise caption (\eg alt-text) describing salient attributes in the image \cite{radford-clip, laion, sharma-2018-conceptualcap}. During training, VLMs generally model the relationship between a paired image-text sample as a \textit{one-to-one} mapping: a single embedding of the entire image is contrastively aligned with a single embedding of the entire caption \cite{radford-clip, zhang2020convirt, jia-align}.

However, real-world multimodal datasets, such as those obtained from healthcare settings or product databases, consist of samples that are significantly more complex than standard image-caption pairs \cite{johnson2019mimic, padchest2020, feta2022, deepfashiondata2016}. In particular, real-world image-text samples include text that describes many distinct attributes occurring in fine-grained regions of the paired image. For example, medical images (\eg X-rays) are accompanied by detailed text reports describing a variety of attributes (\eg characteristics of organs, signs of disease) that map to specific regions of the image \cite{johnson2019mimic, padchest2020, roentgen}. We refer to these samples as exhibiting high \textit{pairwise complexity}, since each image-text pair can be decomposed into a large number of fine-grained region-attribute pairings. Figure \ref{fig:header} provides a visual depiction of pairwise complexity. 

Models with knowledge of region-attribute relationships have been shown to exhibit numerous advantages, ranging from improved performance on fine-grained tasks (such as object detection) to improved subgroup robustness \cite{Zhong_2022_CVPR_regionclip,saab2022spatial,rasheed2022objects}. However, the extent to which standard one-to-one VLMs can learn these fine-grained relationships when trained on complex, real-world datasets is not currently well understood\footnote{As a motivating example, a VLM trained on X-rays should learn to link the region of the heart to the attribute ``enlarged heart'' in the report.}. Prior work \cite{Zhong_2022_CVPR_regionclip} has observed that standard one-to-one VLMs often struggle to learn region-attribute relationships\footnote{When CLIP is applied to a region-level object detection task, performance is 40\% lower than simply applying CLIP at the image level \cite{Zhong_2022_CVPR_regionclip}.}, yet the specific effects of training dataset complexity on the ability of a VLM to capture these relationships have not been previously evaluated.

As our first key contribution in this work, we conduct a systematic evaluation of the effects of training dataset complexity on the fine-grained reasoning ability of standard one-to-one VLMs. To this end, we introduce the \textit{pairwise complexity score}, which measures the number of region-attribute pairings that can be composed from an image-text sample. Then, we create a synthetic dataset \textsc{DocMNIST}, where the average pairwise complexity can be directly controlled by altering the number of region-attribute pairs per sample. We use DocMNIST to demonstrate that as the complexity of the training data increases, a standard one-to-one VLM demonstrates a performance drop of 36.9\% on a text $\rightarrow$ region retrieval task and 20.5\% on a region $\rightarrow$ text retrieval task, as measured by R-Precision scores. Our evaluation suggests that standard one-to-one VLMs are not effective at capturing region-attribute relationships when trained on datasets exhibiting high pairwise complexity. 

Our findings demonstrate the need for a VLM that can learn accurate relationships between image regions and textual attributes when trained on complex multimodal datasets. We establish the following two desiderata for such a model. First, fine-grained relationships should be learned without the need for ground-truth region-attribute pairings, which are generally not provided in datasets and are expensive to manually label (particularly in specialized domains like healthcare). Second, representation learning should be performed using standard one-to-one VLMs. Despite their current inability to capture region-attribute relationships from complex datasets, one-to-one VLMs are highly effective \cite{radford-clip}, widely-used, and less computationally expensive than alternative fine-grained learning approaches \cite{filip2021}. 

As our second key contribution, we present \textbf{Vi}sion-\textbf{L}anguage \textbf{L}earning with \textbf{A}ttributes (\name{}), a self-supervised multimodal representation learning approach that satisfies the above desiderata. Our key insight is that providing region-attribute pairs as training data to a standard one-to-one VLM helps improve the fine-grained reasoning ability of the model. \name{} employs a two stage pipeline, as shown in Figure \ref{fig:header}. The first stage involves training a lightweight \textit{mapping model} to decompose image-text samples into region-attribute pairs. Here, we model the relationship between a paired image-text sample as a \textit{many-to-many} mapping; given a set of many candidate image regions and a set of many textual attributes, we leverage self-supervision to learn a mapping between these sets. Then, the second stage involves training a standard one-to-one VLM on the generated region-attribute mappings. 

We demonstrate with four multimodal training datasets across various domains (synthetic images, product data, medical images, and natural images) that \name{} outperforms comparable one-to-one VLMs on fine-grained reasoning tasks, such as zero-shot object detection (up to 3.6 AP50 points on COCO and 0.6 mAP points on LVIS), text$\rightarrow$region retrieval (up to 14.2 R-Precision points), and region$\rightarrow$text retrieval (up to 7.8 R-Precision points). We show that these improvements are a result of our region-attribute mappings, which are up to 25.8 points more accurate than prior approaches. 

Our contributions are summarized below: 
\begin{itemize}
\item We demonstrate through a series of systematic evaluations that standard one-to-one VLMs struggle to learn relationships between image regions and textual attributes as training dataset complexity increases (leading to performance degradations of up to 37\% on retrieval tasks). We also introduce \textsc{DocMNIST}, a synthetic, customizable training dataset that we hope will be useful for future research on VLMs. 
\item We present \name{}, a self-supervised multimodal representation learning approach that can effectively learn fine-grained region-attribute relationships, particularly when training datasets exhibit high pairwise complexity. We demonstrate that our approach works effectively across a variety of real-world domains, outperforming comparable methods across tasks including zero-shot object detection (COCO and LVIS) and retrieval (CheXpert 5x200). 
\end{itemize}

The rest of this paper is organized as follows. In Section \ref{sec:relatedwork}, we discuss related work. We formally introduce the problem setting in Section \ref{sec:preliminaries}, followed by our analysis with \textsc{DocMNIST} in Section \ref{sec:understanding}. In Section \ref{sec:ourapproach}, we introduce \name{}, and in Section \ref{sec:experiments}, we present experimental results. Finally, we conclude in Section \ref{sec:conclusion}.

\section{Related Work}
\label{sec:relatedwork}
Our work builds on several recent research directions. An extended discussion is provided in Appendix Section \ref{appendix:relatedwork}. 

\smallskip
\noindent
\textbf{One-to-One VLMs:} Recent works have applied contrastive self-supervised learning methods to multimodal datasets \cite{radford-clip,jia-align,pham2021basic,zhang2020convirt,cyclip2022}. During training, each image is pulled towards an associated caption and pushed away from dissimilar captions in the latent space.
\smallskip

\noindent
\textbf{Fine-Grained Representation Learning:} Prior work has shown that one-to-one VLMs often struggle to capture fine-grained region-level information \cite{ma2023crepe}. In particular, \cite{Zhong_2022_CVPR_regionclip} shows that CLIP, a widely-used one-to-one VLM, achieves 60\% accuracy on an image-level classification task yet only 19\% on a region-level classification task with a similar number of classes. The authors attribute the performance drop to the fact that CLIP does not capture fine-grained relationships between image regions and textual attributes. Additionally, \cite{krojer2022imagecode} demonstrates that CLIP often fails to understand subtle differences between images. Our work extends these lines of research by systematically evaluating the effect of training dataset complexity on the fine-grained reasoning ability of one-to-one VLMs. 

Several prior approaches have been proposed for learning fine-grained region-level information from image-text datasets. One line of recent work leverages large quantities of human-labeled region-text pairs during training \cite{glip2022,xvlm,sohn2020,xu2021}. However, obtaining human-annotated region-text pairs is expensive, time-consuming, and difficult to extend to other domains. In order to mitigate the need for human-annotated region-text pairs, \cite{Zhong_2022_CVPR_regionclip} proposes RegionCLIP, which uses the pretrained CLIP model \cite{radford-clip} in a zero-shot fashion to match candidate image regions with plausible textual attributes. However, this approach relies heavily on the CLIP model, which (a) has been shown to work poorly on localizing regions to text \cite{Zhong_2022_CVPR_regionclip} and (b) cannot be accurately applied in a zero-shot fashion to out-of-domain data (such as medical images) \cite{radford-clip}. Here, \name{} aims to address these issues by introducing a specific training phase to learn region-attribute mappings, rather than directly using an off-the-shelf pretrained VLM model. Our work is also inspired by open vocabulary object detection \cite{zareian2021_ovr, vldet, vild2022, rasheed2022objects} and self-supervised patch-token alignment \cite{filip2021,loupe2022,huang2021gloria} methods.

\smallskip
\noindent
\textbf{Learning from Real-World Multimodal Data:} Our work relates closely to prior studies that have developed VLMs for medical \cite{zhang2020convirt, biomedclip,biovil, huang2021gloria} and product datasets \cite{feta2022,Chia2022fashionclip,armani}. We extend these lines of research by developing an approach that can effectively learn fine-grained signal from datasets with high pairwise complexity. We show that \name{} works effectively across multiple real-world domains.

\section{Preliminaries}
\label{sec:preliminaries}

In this section, we formally describe our problem setting. Datasets used for training VLMs can be expressed in the form $\mathcal{D} = \{(x_i, t_i)\}_{i=1}^n$, where $x \in \mathcal{X}$ represents image inputs and $t \in \mathcal{T}$ represents text.  $\mathcal{T}$ often takes the form of concise captions, which are simple phrases describing salient attributes in the associated image. Standard VLMs learn a \textit{one-to-one} alignment between images and captions, which involves learning an embedding function $\psi_{img}: \mathcal{X} \rightarrow R^d$ that maps input images $\mathcal{X}$ to a latent space with dimension $d$. The function $\psi_{img}$ is learned jointly with a function $\psi_{txt}: \mathcal{T} \rightarrow R^d$ that maps text data $\mathcal{T}$ to the same latent space. Current state-of-the-art approaches learn $\psi_{img}$ and $\psi_{txt}$ in a self-supervised manner by leveraging contrastive learning \cite{radford-clip,zhang2020convirt,jia-align,pham2021basic}. 

In this work, we observe that real-world sources of multimodal data often include image-text pairs ($x_i, t_i$) that are complex, where the text refers to a large collection of attributes in various regions of the accompanying image. We can express this formally by decomposing each image into $r_i$ regions, expressed as $x_i = \{x_i^0, x_i^1,…,x_i^{r_i}\}$. Similarly, we decompose each textual description into $a_i$ attributes, expressed as $t_i = \{t_i^0, t_i^1, …, t_i^{a_i}\}$. We note that $r_i$ does not necessarily equal $a_i$, since each textual attribute may manifest in one or more image regions. A set of fine-grained region-attribute pairs of size $m_i$ can be obtained from the original sample pair $(x_i, t_i)$; we refer to $m_i$ as the image-text \textit{pairwise complexity score}. 

Given these definitions, we can quantitatively express the average pairwise complexity of a dataset as $s = \frac{1}{n} \sum_{i=1}^n m_i$, which characterizes the average number of region-attribute pairs per sample. Complex datasets have large values of $s$. Our goal in this work is to introduce an approach that can accurately learn fine-grained relationships between regions and attributes from training datasets with high pairwise complexity.  

In the following section (Section \ref{sec:understanding}), we first demonstrate that as the complexity of a dataset increases, standard one-to-one VLMs struggle to learn fine-grained representations. Then, in Section \ref{sec:ourapproach}, we present an approach to improve vision-language representation learning on datasets exhibiting high pairwise complexity.

\section{Understanding Dataset Complexity}
\label{sec:understanding}
In this section, we aim to better understand the challenges associated with using standard one-to-one VLMs to learn from datasets exhibiting high pairwise complexity. We introduce a synthetic training dataset in Section \ref{sec:docmnist} with a variety of controllable dataset-level properties, such as the number of textual attributes. Then, in Section \ref{sec:docmnisteval}, we use this dataset to demonstrate that as complexity increases, representations learned using standard one-to-one VLMs degrade in fine-grained reasoning ability. 

\subsection{DocMNIST: A synthetic training dataset with controllable pairwise complexity}
\label{sec:docmnist}

Here, we introduce our synthetic vision-language training dataset \textsc{DocMNIST}, which is an adaptation of the popular MNIST benchmark \cite{deng2012mnist}. The purpose of \textsc{DocMNIST} is to enable systematic, controlled evaluations of learned vision-language representations as various dataset-level properties, such as the average pairwise complexity, are modified.

\begin{figure}[t]
\begin{center}
\includegraphics[width=0.99\linewidth]{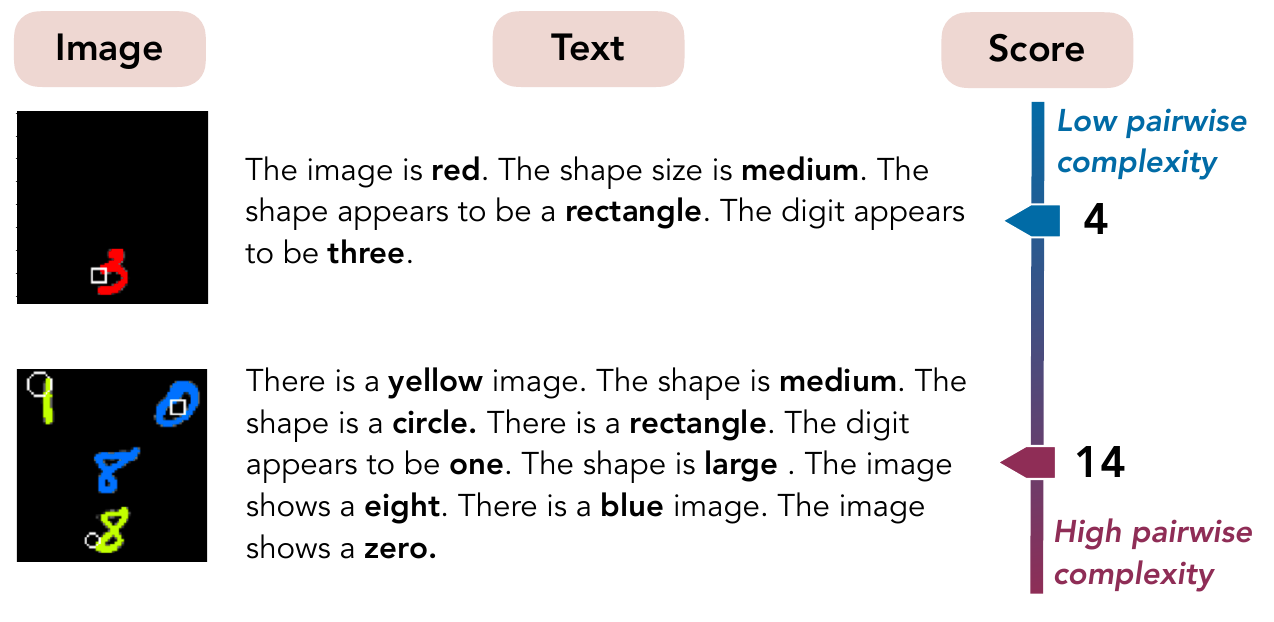}
\end{center}
  \caption{Example image-text pairs from DocMNIST, where each textual attribute (bolded) appears in at least one region of the image. The pairwise complexity score measures the number of distinct region-attribute pairs.}
\label{fig:docmnist}
\end{figure}

\textsc{DocMNIST} consists of images paired with textual descriptions (as shown in Figure \ref{fig:docmnist}). We set the size of each image to be $3 \times 84 \times 84$, subdivided into 9 square regions of size $3 \times 28 \times 28$. We define $A$ as the set of possible attributes that can be assigned to each region. We consider the following attributes ($|A|=20$) across 4 categories: digits (0-9), digit colors (purple, blue, green, yellow, red), shapes (rectangle, circle), and shape sizes (small, medium, large). 

To create a \textsc{DocMNIST} training dataset, we first generate an image by randomly assigning attributes from set $A$ to the nine image regions (\eg a \textit{red six}, which includes two attributes, may be assigned to the top left region). Additional constraints for this assignment process are discussed in further detail in Appendix Section \ref{appendix:docmnist} (\eg all digits must have an associated color). An associated textual description is automatically generated for the image by filling attributes into pre-defined templates (\eg ``The image shows a six."). The average number of region-attribute pairs per sample is controlled by a user-specified variable $c$, which defines the average pairwise complexity of the dataset. 

The final size of the training dataset is constrained by a pre-defined attribute budget $b$, which represents the total number of attributes across all images. We continue generating image-text pairs until the budget $b$ is reached.

In summary, the following dataset-level variables can be controlled when generating a \textsc{DocMNIST} training dataset: the set of possible attributes $A$, the attribute budget $b$, and the average pairwise complexity score $c$. 

\subsection{Evaluating one-to-one VLMs with DocMNIST}
\label{sec:docmnisteval}
In this section, we explore the following key question: \textit{How does the complexity of the training dataset influence the fine-grained reasoning ability of a one-to-one VLM?} We measure fine-grained reasoning ability by determining if learned representations can be used to accurately localize textual attributes to corresponding image regions (text $\rightarrow$ region retrieval) and image regions to corresponding textual attributes (region $\rightarrow$ text retrieval). 

First, we generate a set of \textsc{DocMNIST} training datasets that vary in complexity. We define $A$ as the set of twenty attributes discussed in Section \ref{sec:docmnist}, and we then generate six versions of the training dataset with $c$ ranging from $5.0$ to $29.4$. A training dataset with $c=29.4$ has approximately $24$ more region-attribute pairs per sample than a training dataset with $c=5.0$. A fixed attribute budget of $b = 300K$ is maintained, which ensures that the total number of attributes in each of the six training datasets remains constant.

For each training dataset, we use a one-to-one VLM to contrastively learn alignments between images and the associated text. The image embedding function $\psi_{img}$ is learned using a ResNet-50 model initialized with pretrained CLIP weights \cite{radford-clip, resnet2015}. The text embedding function $\psi_{txt}$ is learned using a CLIP text encoder with frozen weights. 

\begin{figure}[t]
\begin{center}
\includegraphics[width=0.99\linewidth]{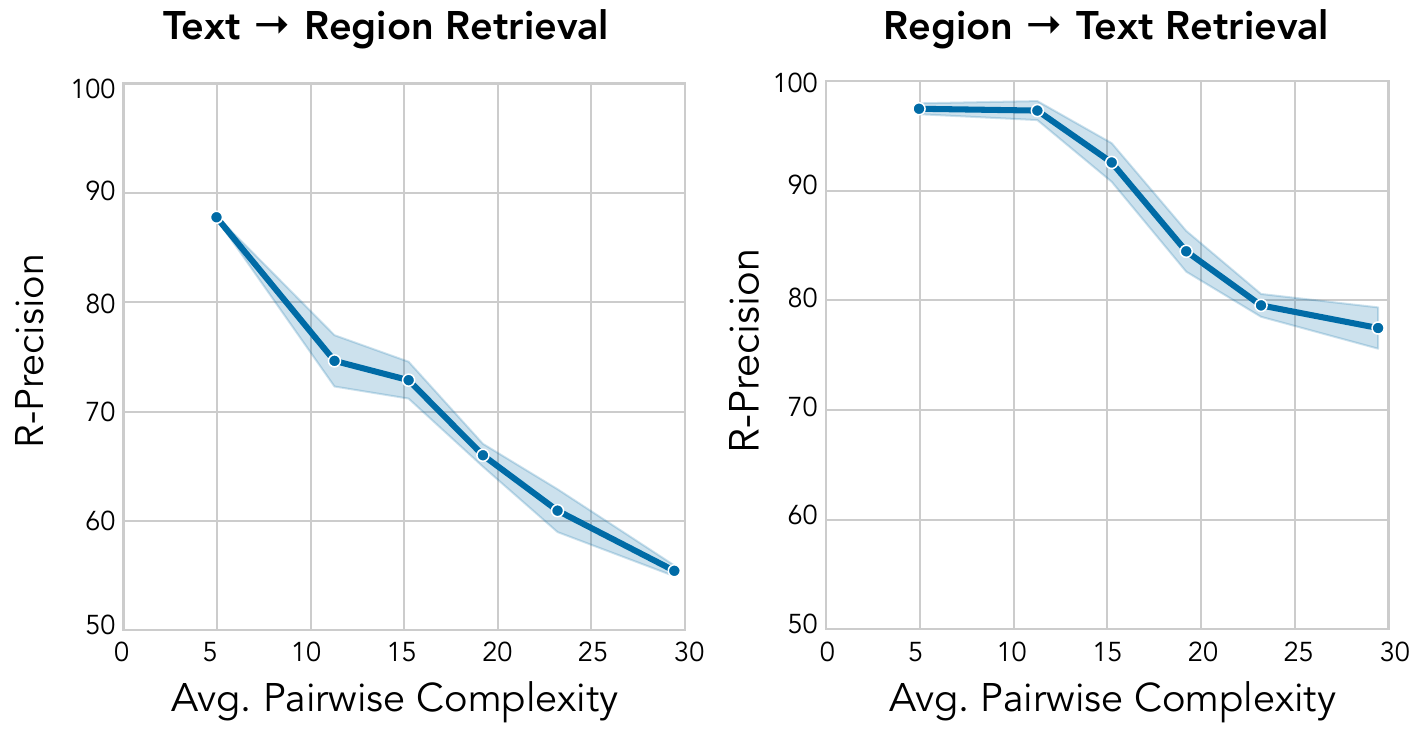}
\end{center}
   \caption{As the complexity of the dataset increases, representations generated using one-to-one VLMs demonstrate lower performance on text $\rightarrow$ region retrieval (left) and region $\rightarrow$ text retrieval (right). We report R-Precision scores.}
\label{fig:docmnistresults}
\end{figure}

We measure the fine-grained reasoning ability of the resulting representations by using a held-out test set to measure text $\rightarrow$ region retrieval and region $\rightarrow$ text retrieval performance. Results are summarized in Figure \ref{fig:docmnistresults}. We observe that text $\rightarrow$ region retrieval performance drops by 36.9\% (32.4 R-Precision points) and region $\rightarrow$ text retrieval performance drops by 20.5\% (20.0 R-Precision points) as average pairwise complexity ($c$) increases from $5.0$ to $29.4$.

In summary, our analysis demonstrates that as the average pairwise complexity of a dataset increases, standard VLMs that assume a one-to-one relationship between images and text struggle to learn fine-grained representations. Additional analysis is provided in Appendix Section \ref{appendix:docmnist}.

\section{Our Approach: \name{}}
\label{sec:ourapproach}
Given our findings from Section \ref{sec:understanding}, we establish the following desiderata as a set of ideal characteristics for a VLM:
\begin{itemize}
    \item A VLM should have knowledge of fine-grained relationships between image regions and textual attributes. These relationships should be learned without supervision, since ground-truth region-attribute pairings are generally not labeled in datasets. 
    \item Standard one-to-one VLMs should be used for learning representations. These models have been shown to be highly effective across a range of tasks \cite{radford-clip}, are widely used across numerous applications, and are less computationally expensive to train than previously-proposed fine-grained learning methods \cite{filip2021}.  
\end{itemize}

In order to satisfy both desiderata, we introduce \name{}, a self-supervised multimodal representation learning approach that uses standard one-to-one VLMs to learn fine-grained region-attribute relationships from complex datasets. \name{} follows a two-stage pipeline: 
\begin{enumerate}
    \item \textit{Stage 1: Mapping Image Regions to Attributes} (Section \ref{subsec:mapping}): Given a set of candidate image regions and textual attributes, we first introduce a self-supervised \textit{mapping model} to generate region-attribute pairs.  
    \item \textit{Stage 2: Learning Vision-Language Representations} (Section \ref{subsec:train}): We then use the generated region-attribute mappings as training data for a standard one-to-one VLM. Our key insight is that providing accurate region-attribute pairs during training will improve the fine-grained reasoning ability of the VLM.
\end{enumerate}

\subsection{Mapping Image Regions to Textual Attributes}
\label{subsec:mapping}
In this section, we discuss Stage 1 of ViLLA, which involves (a) decomposing image-text samples into image regions and textual attributes, (b) constructing a mapping model that learns attribute-specific projections for each input region, and (c) training the mapping model to pair regions and attributes in a self-supervised fashion. We begin with a training dataset $\mathcal{D} = \{(x_i, t_i)\}_{i=1}^n$ consisting of $n$ image-text pairs ($x_i, t_i$). \smallskip

\noindent
\textbf{Decomposing Images and Text:} We model the relationship between each paired image-text sample $(x_i, t_i)$ as a \textit{many-to-many} mapping. We first decompose each image $x_i$ into $r_i$ regions, expressed as $x_i = \{x_i^0, x_i^1,…,x_i^{r_i}\}$. Candidate regions can be identified in several different ways, such as by using off-the-shelf region proposal networks (RPNs), dividing images into equal-sized segments (such as quadrants), or using random bounding boxes. Similarly, we decompose each textual description $t_i$ into $a_i$ attributes, expressed as $t_i = \{t_i^0, t_i^1, …, t_i^{a_i}\}$. Attributes can also be extracted in multiple different ways, such as by leveraging structured labels provided with the dataset or using off-the-shelf entity extraction tools. Optimal approaches for region and attribute selection are dependent on the composition of the training data (\eg whether domain-specific RPNs are available), and we discuss dataset-specific implementation details in Appendix Section \ref{appendix:villaimplementation}. Importantly, \name{} operates under the assumption that each attribute in $t_i$ occurs in at least one region in $x_i$.
\smallskip

\noindent
\textbf{Constructing the Mapping Model:} Given image regions $\{x_i^0, x_i^1,…,x_i^{r_i}\}$ and textual attributes $\{t_i^0, t_i^1, …, t_i^{a_i}\}$, our goal is to learn mappings between regions and attributes. The mapping model consists of an image encoder to generate embeddings for image regions and a text encoder to generate embeddings for textual attributes. 

In order to generate embeddings for image regions, we begin by initializing a CNN with pretrained weights from a domain-specific one-to-one VLM; we use ConVIRT weights \cite{zhang2020convirt} for our medical dataset and CLIP-RN50 weights \cite{radford-clip} for all non-medical datasets. Given image $x_i$ as input, we use RoIAlign \cite{maskrcnn-roialign, Zhong_2022_CVPR_regionclip} to extract a representation for each region, resulting in region embeddings expressed in matrix form as $e_i \in \mathbb{R}^{r_i \times d}$. We use the variable $d$ to represent the output embedding dimension, which is set to 1024 for the CLIP-based models and 768 for the ConVIRT-based model. Then, region embeddings $e_i$ are provided as input to a set of $p$ projection heads. Each projection head consists of a linear layer, a ReLU function, and a second linear layer. In most cases, we set $p = |A|$, where $A$ represents the set of all attributes in the input dataset; as a result, each projection head is associated with a single textual attribute and yields an embedding that characterizes regions with respect to the attribute\footnote{In some cases, particularly when $|A|$ is large, we set $p < |A|$ and assign each projection head to multiple textual attributes. Analysis on the selection of $p$ is provided in Appendix Section \ref{appendix:villaimplementation}.}. We represent the projection head for attribute $k$ with the function $P_k$, which outputs final region embeddings of size $r_i \times d$. 

Next, in order to generate embeddings for textual attributes, we insert each attribute into pre-defined prompt templates (\eg ``a photo of an [attribute]"). We extract representations from a pretrained text encoder, which is initialized with weights from sentence-BERT (SBERT) \cite{reimers-2019-sbert} for our medical dataset and CLIP for all non-medical datasets \cite{radford-clip}. This yields an embedding $h_i \in \mathbb{R}^{a_i \times d}$ for textual description $t_i$. Additional implementation details are provided in Appendix Section \ref{appendix:villaimplementation}.  
\smallskip

\noindent
\textbf{Training Procedure:} Given region embeddings $e_i$, projection heads $P$, and attribute embeddings $h_i$, we now use the following procedure to train the mapping model. In order to ensure that the mapping model is lightweight, we freeze all parameters in the image and text encoder, leaving only the projection head parameters as trainable. Given a textual attribute $k \in t_i$, we use the notation $P_k(e_i) \in \mathbb{R}^{r_i \times d}$ to refer to the region embeddings resulting from the projection head corresponding to attribute $k$. Similarly, we use the notation $h_i^k \in \mathbb{R}^{1 \times d}$ to refer to an embedding of textual attribute $k$.

We train the mapping model with the following contrastive loss function. Let $B$ represent the batch (with $|B|$ image-text pairs) and let $\sigma(a,b) = \exp(\max(\langle a, b \rangle / \tau))$. 
\begingroup\makeatletter\def\f@size{8.5}\check@mathfonts
$$L(x_i, t_i) = -\sum_{k \in t_i}  \log \frac{\sigma(P_k(e_i), h_i^k)}{\sigma(P_k(e_i), h_i^k) + \sum_{j=1; {k \not\in t_j}}^{|B|}\sigma(P_k(e_j), h_i^k)}$$\endgroup

For sample $(x_i, t_i$) and attribute $k \in t_i$, this loss function encourages the maximum pairwise similarity between region embeddings $P_k(e_i)$ and the attribute embedding $h_i^k$ to be high, since at least one region in $x_i$ depicts attribute $k$. Simultaneously, for an image $x_j$ where $k \not\in t_j$, the maximum pairwise similarity between the region embeddings $P_k(e_j)$ and the attribute embedding $h_i^k$ is encouraged to be low, since no regions in $x_j$ depict attribute $k$.

\subsection{Learning Vision-Language Representations}
\label{subsec:train}
In this section, we discuss Stage 2 of ViLLA, which involves computing region-attribute pairs based on similarity scores assigned by the mapping model; then, a one-to-one VLM is trained on the generated pairs.

The mapping model from Section \ref{subsec:mapping} can be used to assign attributes to regions as follows. For a sample ($x_i, t_i$) and textual attribute $k \in t_i$, we compute the pairwise dot product between $P_k(e_i)$ and $h_i^k$, resulting in a score vector $v \in \mathbb{R}^{r_i \times 1}$. We then assign $k$ to all regions with a score greater than $max(v) - \epsilon$, where $\epsilon$ is a pre-defined threshold. This could assign zero or multiple attributes to a region\footnote{There are some settings, such as when regions are tight bounding boxes from an RPN, where each region is likely to capture exactly one attribute. In these cases, we invert this process and instead assign regions to attributes. Details are provided in Appendix Section \ref{appendix:villaimplementation}.}.

We then augment the training dataset to include generated region-attribute pairs in addition to the original image-text samples. We use the augmented dataset to train a one-to-one VLM, which is optimized using a standard bidirectional contrastive loss function \cite{zhang2020convirt, radford-clip}. 

\section{Experiments}
\label{sec:experiments}
We evaluate our approach using four training datasets from various domains (synthetic images, product data, medical images, and natural images) and three fine-grained reasoning tasks (zero-shot object detection, text $\rightarrow$ region retrieval, region $\rightarrow$ text retrieval). Our experiments show that (1) our approach outperforms prior methods across all three tasks (Section \ref{sec:downstream}) and (2) our region-attribute mappings are more accurate than prior approaches (Section \ref{sec:mappingaccresults}). We provide extended results in Appendix Section \ref{appendixsec:eval}.

\subsection{Datasets}
We apply \name{} to learn vision-language representations from four training datasets: DocMNIST (synthetic images), DeepFashion (product data), MIMIC-CXR (medical images), and COCO (natural images). Table \ref{tab:dataset} includes summary statistics, and further details are provided below:

\begin{table}
\begin{center}
\begin{tabular}{lcccc}
\hline
Dataset & Domain & Regions  & Attributes & $s$ \\
\hline
DocMNIST   & Synthetic & 8.9 & 20 & 29.4 \\
DeepFashion & Product & 4 & 58 & 7.9\\
MIMIC-CXR   &  Medical & 3 & 50 & 5.0  \\
COCO   &  Natural & 300 & 4.7k & 6.8\\
\hline
\end{tabular}
\end{center}
\caption{We use \name{} to learn representations from 4 multimodal training datasets. Datasets vary in the average number of regions per image and total number of attributes. We also estimate the average pairwise complexity score ($s$) of each dataset.}
\label{tab:dataset}
\end{table}

\smallskip
\noindent
\textbf{DocMNIST}: We create the synthetic DocMNIST dataset using the procedure described in Section \ref{sec:understanding}. We generate a training dataset with an average pairwise complexity of 29.4.

\smallskip
\noindent
\textbf{DeepFashion} \cite{deepfashiondata2016,jiang2022text2human}: The DeepFashion-MultiModal dataset consists of 44k images extracted from clothing retail websites. Each image is accompanied with multi-sentence textual descriptions and structured labels (\eg sleeve length, hats, etc.). During training, we use the 58 provided structured labels as our set of relevant attributes, and we divide each image lengthwise into 4 regions. 

\smallskip
\noindent
\textbf{MIMIC-CXR} \cite{johnson2019mimic, physionet}: The MIMIC-CXR dataset consists of 377k chest X-ray images and associated physician reports obtained from the Beth Israel Deaconess Medical Center. We train an anatomy-specific RPN to divide each image into 3 regions: right lung, left lung, and heart. In order to create our attribute set, we use an off-the-shelf entity extractor (RadGraph) to identify the 50 entities that occur most frequently in the reports \cite{radgraph}.

\smallskip
\noindent
\textbf{COCO} \cite{coco2014}: The Microsoft COCO training dataset consists of 114k natural images. Each image is associated with five captions. In line with prior work, we extract 4.7k textual attributes (\eg giraffe, man, bicycle, etc.) from the captions \cite{Zhong_2022_CVPR_regionclip}. For each image, we use a pretrained RPN to extract 300 candidate regions \cite{fasterrcnn}.

\subsection{Downstream Task Evaluations}
\label{sec:downstream}

We evaluate \name{} on three fine-grained reasoning tasks: (1) zero-shot object detection, (2) text $\rightarrow$ region retrieval, and (3) region $\rightarrow$ text retrieval. 

\subsubsection{Zero-Shot Object Detection}
\label{sec:objectdetection}

\begin{table*}[t]
\begin{center}
\begin{tabular}{lcccccccccc}
\toprule
\multicolumn{2}{c}{} & \multicolumn{4}{c}{COCO} & \multicolumn{1}{c}{} & \multicolumn{4}{c}{LVIS} \\ 
\cmidrule{3-6}
\cmidrule{8-11}

Method & Pretraining Dataset  & AP\textsubscript{small} & AP50\textsubscript{Novel} & AP50\textsubscript{Base} & AP50\textsubscript{All} & & APr & APc & APf & mAP\\
\midrule
OVR-CNN  & COCO  & - & 46.7 & 43.7 & 44.5 & & - & - & - & - \\
CLIP  & CLIP400M  & 43.3 & 58.6 & 58.2 & 58.3  & & 40.3 & 41.7 &  43.6  & 42.2\\
RegionCLIP  & CC3M  & 47.3 & 60.5 & 61.7 & 61.4 & & 40.7 & \textbf{43.5} & 47.0 & 44.4\\
RegionCLIP  & COCO  & - & - & - & 62.8 &  &  - & - & - & -\\
\name{} (Ours)  & COCO  & \textbf{51.9} & \textbf{63.5} & \textbf{67.4} & \textbf{66.4} & & \textbf{42.6} & 42.4 & \textbf{49.0} & \textbf{45.0} \\

\bottomrule
\end{tabular}
\end{center}
\caption{\textit{Zero-shot object detection results.} We compare our approach with prior zero-shot object-detection methods. We report average precision (AP) scores on COCO (small, novel, base, and all object classes) and LVIS (rare, common, frequent, and all object classes).}
\label{tab:obj}

\end{table*}

\begin{table*}[t]
\begin{center}
\begin{tabular}{lcccccccccc}
\toprule
\multicolumn{1}{c}{} & \multicolumn{7}{c}{Text $\rightarrow$ Region} & \multicolumn{1}{c}{} & \multicolumn{2}{c}{Region $\rightarrow$ Text} \\ 
\cmidrule{2-8}
\cmidrule{10-11}
\multicolumn{1}{c}{} & \multicolumn{3}{c}{DocMNIST} & \multicolumn{1}{c}{} & \multicolumn{3}{c}{DeepFashion} & \multicolumn{1}{c}{} & \multicolumn{1}{c}{DocMNIST} & \multicolumn{1}{c}{DeepFashion} \\ 
Method & P@25 & P@100 & R-Prec & & P@25 & P@100 & R-Prec & & R-Prec & R-Prec\\
\midrule
\textrm{CLIP-ZS} & 46.4 & 42.1 & 30.2 & & 30.6 & 31.8 & 19.0 & & 54.2 & 10.9 \\
\textrm{CLIP-FT-Img} & 84.4 & 78.2 & 55.2 & & 51.6 & 58.0 & 34.4 & & 78.7 & 27.3 \\
\textrm{CLIP-FT-Reg} & 47.0 & 38.8 & 29.0 & & 33.6 & 36.3 & 23.8 & & 38.4 & 21.0 \\
\textrm{CLIP-ZS-Map} & 55.6 & 47.6 & 35.2 & & 53.9 & 59.4 & 35.3 & & 49.9 & 28.8 \\ 
\textrm{\name{} (Ours)} & \textbf{91.6} & \textbf{91.6} & \textbf{69.4} & & \textbf{56.5} & \textbf{67.4} & \textbf{39.5} & & \textbf{86.5} & \textbf{32.3}  \\
\bottomrule
\end{tabular}
\end{center}
\caption{\textit{Retrieval on DocMNIST and DeepFashion:} We report text $\rightarrow$ region and region $\rightarrow$ text retrieval results on the DocMNIST and DeepFashion datasets. We compare our approach with several CLIP-based baselines, which ablate various components of our method.}
\label{tab:ret}
\end{table*}

\begin{table}[t]
\begin{center}
\begin{tabular}{lcccc}
\toprule
Method  &  One-to-One & Accuracy \\
\midrule
ConVIRT    & \cmark &   49.0  \\
BioViL    & \cmark & 47.6    \\
GLoRIA-Global Only   & \xmark &  52.1   \\
GLoRIA-Local Only    & \xmark &  41.7  \\
GLoRIA-Global+Local    & \xmark & 48.8     \\
\name{} (Ours)  & \cmark & \textbf{55.9}\\
\bottomrule
\end{tabular}
\end{center}
\caption{\textit{Retrieval on CheXpert 5x200}: We compare \name{} with prior models, only two of which are one-to-one VLMs.}
\label{tab:chexpert}
\end{table}

\noindent
\textbf{Task:} Given a bounding box containing an object, the zero-shot object detection task involves localizing the bounding box to an object category without performing any task-specific fine-tuning. We evaluate performance on the COCO validation dataset, which consists of 4.8k images annotated with ground-truth object bounding boxes corresponding to 65 classes \cite{coco2014}. We also evaluate on the LVIS validation set, which consists of 19k images across 1000 classes \cite{gupta2019lvis}. We set up both tasks as described in \cite{Zhong_2022_CVPR_regionclip}.  

\smallskip
\noindent
 \textbf{Evaluation:} We train \name{} on the COCO training dataset. We use a previously-developed evaluation framework \cite{Zhong_2022_CVPR_regionclip} to compare our learned representations against three prior methods: OVR-CNN \cite{zareian2021_ovr}, CLIP \cite{radford-clip}, and RegionCLIP \cite{Zhong_2022_CVPR_regionclip}. In line with prior work, we report AP50 scores on COCO across 17 novel categories, 48 base categories, and all 65 categories; we also report AP\textsubscript{small} in order to characterize performance on small objects, a particularly challenging subgroup. On LVIS, we report AP across 337 rare categories (APr), 866 common and frequent categories (APc and APf), and all categories (mAP). 

\smallskip
\noindent
\textbf{Results:} Results are in Table \ref{tab:obj}. On COCO, our approach contributes to 8.1 points of improvement in AP50 over CLIP and 3.6 points of improvement over RegionCLIP. On the challenging subgroup of small objects, we note improvements of 8.6 points over CLIP and 4.6 points over RegionCLIP, suggesting that \name{} is accurate even when regions are small. Similarly, on LVIS, we observe 2.8 points of improvement in mAP over CLIP and 0.6 points of improvement over RegionCLIP. Our results indicate that \name{} is able to effectively reason over region-attribute relationships.

\subsubsection{Text $\rightarrow$ Region Retrieval}
\label{sec:textregion}
\noindent
\textbf{Task:} Given a textual query (\eg ``The person is wearing a hat"), the text $\rightarrow$ region retrieval task determines if we can retrieve image regions that capture the content of the query. We evaluate text $\rightarrow$ region retrieval on a held-out DocMNIST test set consisting of 5.9k regions as well as a held-out DeepFashion test set consisting of 7.9k regions. We obtain textual queries for both datasets by inserting attributes into pre-defined prompt templates. We consider 20 queries for DocMNIST and 46 queries for DeepFashion. 

\smallskip
\noindent
\textbf{Evaluation:} We train \name{} on the DocMNIST and DeepFashion training datasets. We compare our representations against four baselines: (1) CLIP-ZS, which applies CLIP to this task in a zero-shot manner, (2) CLIP-FT-Img, which applies a CLIP model fine-tuned on image-text pairs, (3) CLIP-FT-Reg, which applies a CLIP model fine-tuned by aligning each region to the entire textual description, and (4) CLIP-ZS-Map, which first uses CLIP to generate region-attribute pairs in a zero-shot manner and then fine-tunes a CLIP model with the generated samples. We note that CLIP-ZS-Map is comparable to the RegionCLIP approach explored in Section \ref{sec:objectdetection}. We report Precision@25, Precision@100, and R-Precision. 

\smallskip
\noindent
\textbf{Results:} Results are summarized in Table \ref{tab:ret}. On DocMNIST, our approach contributes to 13.4 points of improvement in P@100 and 14.2 points of improvement in R-Precision over CLIP-FT-Img, which is the next highest baseline. Similarly, on DeepFashion, our approach contributes to 8.0 points of improvement in P@100 and 4.2 points of improvement in R-Precision over CLIP-ZS-Map, which is the next highest baseline. In particular, we note that CLIP-ZS-Map, which is a baseline that emulates the design of RegionCLIP, performs particularly poorly on the DocMNIST dataset since the generated region-attribute mappings are limited by the performance of the original CLIP model. We also note that our performance improvements on DocMNIST are higher than DeepFashion; this is in line with the statistics presented in Table \ref{tab:dataset}, which indicate that DocMNIST has a higher average pairwise complexity than DeepFashion. Our approach is most effective on datasets with high pairwise complexity scores. 

\subsubsection{Region $\rightarrow$ Text Retrieval}
\label{sec:regiontext}
\textbf{Task:} Given an image region, the region $\rightarrow$ text retrieval task determines if we can identify the textual attributes depicted in the region. Again, we use 5.9k regions and 20 attributes for DocMNIST and 7.9k regions and 46 attributes for DeepFashion.

We additionally evaluate retrieval performance on the CheXpert 5x200 benchmark, which consists of 1000 chest X-rays across five disease categories \cite{huang2021gloria}. Each disease label is converted into text using pre-defined prompts. Then, given a chest X-ray, the goal is to retrieve the textual phrase corresponding to the correct disease. 

\smallskip
\noindent
\textbf{Evaluation:} We train \name{} on the DocMNIST and DeepFashion datasets, and we compare our representations against the four baselines described in Section \ref{sec:textregion}. Here, we report R-Precision scores. 

For the CheXpert 5x200 task, we train \name{} on the MIMIC-CXR dataset. At evaluation time, we consider each X-ray as a set of four regions -  right lung, left lung, heart, full image - and perform retrieval by computing the maximum pairwise similarity with the text phrases. We compare with three prior methods: (1) ConVIRT \cite{zhang2020convirt, vilmedic2022}, (2) GLoRIA \cite{huang2021gloria, vilmedic2022}, and (3) BioViL \cite{biovil}. We report accuracy. 

\smallskip
\noindent
\textbf{Results:} Results on the DocMNIST and DeepFashion retrieval tasks are summarized in Table \ref{tab:ret}. On DocMNIST, our approach contributes to 7.8 points of improvement over CLIP-FT-Img, which achieves the next highest score. We also achieve 3.5 points of improvement on DeepFashion over the next highest score. Again, we note that the improvements on DocMNIST are larger than DeepFashion.

Table \ref{tab:chexpert} shows results on the CheXpert 5x200 task. Our approach contributes to 3.8 points of improvement over GLoRIA, which achieves the next highest score. Two prior methods in Table \ref{tab:chexpert} are one-to-one VLMs; we surpass these models by 6.9 and 8.3 points respectively.  

\subsection{Evaluating Region-Attribute Mappings}
\label{sec:mappingaccresults}

\begin{figure}[t]
\begin{center}
\includegraphics[width=0.99\linewidth]{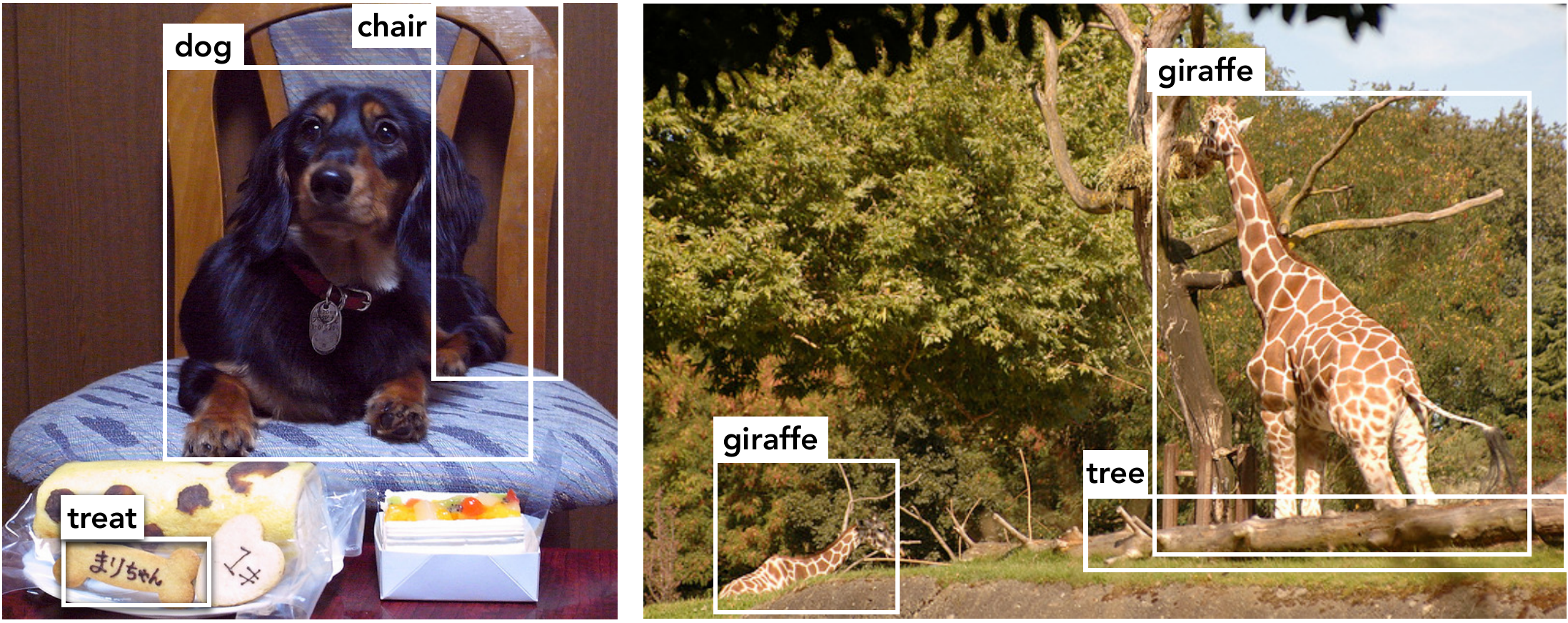}
\end{center}
  \caption{Examples of region-attribute pairs generated by \name{}.}
\label{fig:examples}
\end{figure}

In this section, we demonstrate that the downstream performance improvements observed in Section \ref{sec:downstream} result from the improved quality of our region-attribute mappings. We evaluate the accuracy of region-attribute mappings on a test set associated with each of our pretraining datasets. In Table \ref{tab:mapping_acc}, we compare our approach to a random baseline as well as VLM-ZS, which refers to a one-to-one VLM (CLIP for DocMNIST, DeepFashion, and COCO and ConVirt for MIMIC) applied in a zero-shot manner. The VLM-ZS approach has been used in prior work to map regions to attributes \cite{Zhong_2022_CVPR_regionclip}. Our results demonstrate that our approach outperforms baselines by up to 25.8 F1 points, suggesting that our mappings are higher quality than previous approaches and are contributing to improvements in representation quality. In Figure \ref{fig:examples}, we provide examples of region-attribute pairs generated by \name{} on the COCO dataset. 

\begin{table}[t]
\begin{center}
\begin{tabular}{lccccc}
\toprule
 Method  &  \small DocMNIST & \small DeepFashion & \small MIMIC & \small COCO\\
\midrule
Random  & 27.4 & 32.0 & 35.4 & 38.4\\
VLM-ZS    & 42.6 & 54.8 & 61.8 & 72.9   \\
ViLLA  & \textbf{68.4} & \textbf{74.1} & \textbf{74.5} & \textbf{77.8} \\
\bottomrule
\end{tabular}
\end{center}
\caption{\textit{Mapping quality}: We compare region-attribute pairs generated by \name{} with previous approaches \cite{Zhong_2022_CVPR_regionclip}. VLM-ZS refers to a pretrained one-to-one VLM (CLIP or ConVIRT) applied in a zero-shot manner \cite{radford-clip,zhang2020convirt}. We report F1 scores.}
\label{tab:mapping_acc}
\end{table}

\section{Conclusion}
\label{sec:conclusion}

In this work, we first demonstrate with evaluations on \textsc{DocMNIST} that as the complexity of the training dataset increases, standard one-to-one VLMs struggle to capture fine-grained region-attribute relationships. To address this issue, we introduce \name{}. We demonstrate through evaluations with multiple real-world datasets that \name{} can effectively capture region-attribute relationships, even when training datasets exhibit high pairwise complexity. Limitations of our work include: (a) our evaluations are currently limited to image-text datasets, and (b) our evaluation of region-attribute mapping accuracy is limited on datasets like MIMIC-CXR that do not include ground-truth annotations. Future directions include extending our approach to real-world datasets with other data modalities (such as audio, video, time-series) and conducting user studies to better evaluate the quality of region-attribute mappings on datasets that lack ground-truth annotations.


\section*{Acknowledgments}
MV is supported by graduate fellowship awards from the Department of Defense (NDSEG) and the Knight-Hennessy Scholars program at Stanford University. SH is supported by the Fannie and John Hertz Foundation, the National Science Foundation Graduate Research Fellowship under Grant No. DGE-1656518, and as a Texas Instruments Fellow under the Stanford Graduate Fellowship in Science and Engineering. AC is supported by NIH grants R01 AR077604, R01 EB002524, R01 AR079431, and P41 EB027060; NIH contracts 75N92020C00008 and 75N92020C00021. CL is supported by NIH grants R01 HL155410, R01 HL157235, by AHRQ grant R18HS026886, by the Gordon and Betty Moore Foundation, and by the National Institute of Biomedical Imaging and Bioengineering (NIBIB) under contract 75N92020C00021. 

{\small
\bibliographystyle{ieee_fullname}
\bibliography{paper}
}
\clearpage
\onecolumn
\appendix
\noindent {\Large\textbf{Appendix}}
\tableofcontents

\resumetocwriting
\section{Extended Related Work}
\label{appendix:relatedwork}

\noindent
In this section, we provide an extended analysis of related work. Our work builds on several directions explored in prior studies.

\smallskip
\smallskip
\noindent
\textbf{One-to-One Vision-Language Models:} Early works in computer vision performed image representation learning by training on large labeled datasets, where annotations were primarily obtained from crowdworkers \cite{resnet2015,imagenet2012}. However, the recent development of contrastive self-supervised methods have greatly reduced the need for large-scale, annotated datasets. Driven by methods like SimCLR and MoCo, contrastive self-supervised methods learn representations by pulling similar images together and pushing dissimilar images apart in the latent space \cite{chen2020simclr, moco}. Recent works (\eg CLIP, ALIGN, BASIC, ConVirt) extended these approaches to multimodal datasets, where each image is pulled towards an associated caption and pushed away from dissimilar captions in the latent space \cite{radford-clip,jia-align,pham2021basic,zhang2020convirt,cyclip2022}. When trained over large-scale datasets consisting of millions of image-text pairs collected from the web, these vision-language models (VLMs) were shown to be highly effective across a variety of classification, retrieval, and robustness tasks \cite{domino2022,cliprobustness2022,radford-clip,jia-align,pham2021basic}. We refer to this class of VLMs as one-to-one models, since a single embedding of the entire image is aligned with a single embedding of the textual caption. 

\smallskip
\smallskip
\noindent
\textbf{Fine-Grained Representation Learning:} Models that possess knowledge of fine-grained region-level information have been shown in prior work to exhibit a number of advantages. For example, given the image of the cow shown in Figure \ref{fig:short}, a model that can effectively capture region-level details would learn the features (\eg cow, trees, clouds, etc.) appearing in each bounding box. Computer vision models that can understand details at the region-level are  particularly useful for fine-grained reasoning tasks, such as region-level retrieval and object detection \cite{regioncrossmodal, Zhong_2022_CVPR_regionclip,rasheed2022objects,vldet}. Additionally, \cite{saab2022spatial} demonstrated that medical image models provided with pathology bounding boxes during training are more resistant to spurious correlations, suggesting that knowledge of region-level information can also improve model robustness. These works demonstrate the importance of developing models that can capture region-level details. 

\smallskip
\smallskip
\noindent
However, recent studies have demonstrated that standard one-to-one VLMs often struggle to capture fine-grained region-level information \cite{ma2023crepe}. In particular, \cite{Zhong_2022_CVPR_regionclip} applied CLIP, a widely-used one-to-one VLM, to both an image classification task (ImageNet \cite{imagenetdataset2009}) and a region classification task (LVIS \cite{gupta2019lvis}) with a similar number of classes. Results showed that classification accuracy dropped significantly from 60\% on ImageNet to 19\% on LVIS, which the authors hypothesized was likely due to the fact that CLIP was trained in a one-to-one fashion on image-text pairs and did not learn to capture fine-grained relationships between image regions and textual attributes. Additionally, \cite{krojer2022imagecode} presents a distinct but related study demonstrating that one-to-one VLMs like CLIP often fail to understand subtle differences between images. In particular, given a contextual description, VLMs are evaluated on their ability to retrieve the correct image from a set of ten candidate images that vary only in fine-grained details. Here, CLIP performs almost 60 points worse than human accuracy. Our work extends these lines of research by systematically evaluating the effect of image-text pairwise complexity in the training dataset on the fine-grained reasoning ability of one-to-one VLMs. 

\smallskip
\smallskip
\noindent
Several prior approaches have been proposed for learning fine-grained region-level information from image-text datasets. One line of recent work (\eg GLIP, X-VLM) leverages large quantities of human-labeled region-text pairs during model training \cite{glip2022,xvlm}. A similar line of research has used object detectors pretrained on labeled region-text pairs to generate pseudo-labels for image regions in a semi-supervised fashion \cite{sohn2020,xu2021}; the generated region-label pairs can then be used during model training. Although both approaches have led to significant improvements on vision-language tasks, obtaining human-annotated region-text pairs is expensive, time-consuming, and difficult to extend to other domains, such as medical images. In order to mitigate the need for human-annotated region-text pairs, \cite{Zhong_2022_CVPR_regionclip} proposed RegionCLIP, which uses the pretrained CLIP model \cite{radford-clip} in a zero-shot fashion to match candidate image regions with plausible textual attributes; these mappings are then utilized during training. This approach does not require ground-truth region-text pairs and was shown to work well across open vocabulary and zero-shot object detection tasks; however, this approach relies heavily on the CLIP model, which (a) has been shown to work poorly on localizing regions to text \cite{Zhong_2022_CVPR_regionclip} and (b) cannot be accurately applied in a zero-shot fashion to out-of-domain data (such as medical images) \cite{radford-clip}. Our approach \name{} aims to address these issues by introducing a specific training phase to learn region-text mappings, rather than directly using an off-the-shelf pretrained VLM model. Our work is also inspired by other recent studies in open vocabulary object detection \cite{zareian2021_ovr, vldet, vild2022, rasheed2022objects}.

\smallskip
\smallskip
\noindent
Another related line of work aims to learn fine-grained patterns by aligning individual image patches with textual tokens. \cite{filip2021} proposed FiLIP, which leverages self-supervised contrastive learning to learn token-wise similarities between individual image patches (generated from a vision transformer) and text tokens (generated from a text transformer). LOUPE follows a similar set-up, although a game theoretic algorithm is utilized to learn fine-grained interactions \cite{loupe2022}. In the medical domain, \cite{huang2021gloria} introduced GLoRIA, which adds a local contrastive loss term to the optimization objective in order to match image patches with textual tokens. Although these methods demonstrate performance improvements over standard one-to-one VLMs on several tasks, learning relationships between all image patches and all text tokens is extremely computationally expensive (\eg FILIP leverages a number of computational tricks in order to make training feasible) and is likely to model patches and tokens that are not semantically meaningful (\eg solid-color image patches with no discernible features). Furthermore, these approaches will be particularly expensive when applied to image-text pairs with high pairwise complexity, which are likely to have a large number of textual tokens.  

\smallskip
\smallskip
\noindent
\textbf{Learning from Real-World Multimodal Data:} Our work relates closely to prior studies that have developed VLMs for real-world datasets. 

\smallskip
\smallskip
\noindent
In the medical domain, early works learned medical image representations by training classification models with either (a) labels generated by experts, which are often time-consuming and expensive to obtain, or (b) labels obtained from applying labeling rules to radiologist reports, which are often noisy and limited to a few pre-defined categories \cite{Wang2020, Abrmoff2016, chexpert2019, Wang_2017}. In order to address these issues, \cite{zhang2020convirt} proposed ConVIRT, the first contrastive, self-supervised VLM for learning representations directly from chest X-rays and radiologist reports. ConVIRT, which leverages a one-to-one training approach, was shown to significantly outperform image-only baselines across a range of medical tasks. Other one-to-one VLMs (\eg BiomedCLIP, BioViL) have also been introduced for biomedical tasks \cite{biomedclip,biovil}. In order to improve the fine-grained reasoning ability of ConVIRT, \cite{huang2021gloria} proposed GLoRIA, which uses a local contrastive loss function as discussed in the previous section. Adding this extra loss computation is expensive, and the model is trained using only a small portion of the radiology report (“Impression” section). In the product domain, several recent studies have focused on training VLMs on fashion data, IKEA catalogs, and car manuals \cite{feta2022,Chia2022fashionclip}. In order to capture finer-grained signal from multimodal fashion datasets, \cite{armani} introduced MaskCLIP, which aligns image patches obtained from a garment segmentation mask with textual tokens extracted from simple phrase-based captions; MaskCLIP is introduced as part of a pipeline for synthesizing fashion designs and is not directly evaluated on fine-grained reasoning tasks. 

\smallskip
\smallskip
\noindent
Our work extends upon these lines of research by introducing a lightweight training mechanism for capturing fine-grained relationships even when training datasets possess high image-text pairwise complexity. Of particular note, we demonstrate that our approach \name{} can effectively learn representations across multiple real-world domains.

\section{Extended Dataset Complexity Evaluations with DocMNIST}
\label{appendix:docmnist}

\noindent
In this section, we provide additional details on our dataset complexity evaluations as described in Section \ref{sec:understanding}. 

\subsection{DocMNIST Implementation Details}
\label{appendixsec:docmnistimp}
\noindent
We introduce \textsc{DocMNIST} (adapted from the popular MNIST benchmark \cite{deng2012mnist}), a customizable multimodal training dataset consisting of synthetically-generated image-text pairs. 

\smallskip
\smallskip
\noindent
We first discuss our methodology for generating images. Each image in \textsc{DocMNIST} is set to a size of $3 \times 84 \times 84$; the image is then subdivided into a square grid with 9 regions of size $3 \times 28 \times 28$. We establish an attribute set $A$ consisting of 20 attributes divided across 4 categories as follows: 10 digits (zero, one, two, three, four, five, six, seven, eight, nine), 5 digit colors (purple, blue, green, yellow, red), 2 shapes (rectangle, circle), and 3 shape sizes (small, medium, and large). We selected these attributes in order to emulate properties of real-world multimodal training datasets; for instance, size-based attributes are often used in medical reports when describing anatomic features (\eg “the heart appears enlarged”). We use the following procedure to assign attributes to regions:
\begin{itemize}
\item First, we randomly choose a region from the set of nine possible regions.
\item We randomly select a digit from the set of ten possible digits \{zero, one, two, three, four, five, six, seven, eight, nine\}. We then sample a digit image from the MNIST dataset with the selected digit label and resize the image to $3 \times 28 \times 28$.  The digit image is pasted within the chosen region. 
\item We randomly select a color from the set of five possible colors \{purple, blue, green, yellow, red\}. The selected color is applied to the digit. 
\item Next, we randomly sample a shape from the set \{no shape, rectangle, circle\}. If \textit{no shape} is selected, then no further action is taken and the region will consist of only a digit and a color. If a rectangle or circle is selected, we then randomly sample a size from the set \{small, medium, large\}. The shape is drawn in a random location within the region. For circles, \textit{small} corresponds to a radius of 1 pixel, \textit{medium} corresponds to a radius of 3 pixels, and \textit{large} corresponds to a radius of 5 pixels. For rectangles, \textit{small} corresponds to a length and width of 1 pixel, \textit{medium} corresponds to a length and width of 4 pixels, and \textit{large} corresponds to a length and width of 7 pixels.
\end{itemize}
For each image, we repeat the above procedure until the number of region-attribute pairs reaches the user-specified value for $c$, which defines the average pairwise complexity. The final size of the training dataset is constrained by a pre-defined attribute budget $b$, which represents the total number of attributes across all images. We continue generating image-text pairs until the budget $b$ is reached.

\smallskip
\smallskip
\noindent
Next, we discuss our methodology for generating text. For each image, we automatically generate a textual description by filling the selected attributes into pre-defined templates. Templates are defined for each attribute category as follows, where components in square brackets are replaced with the corresponding attribute:
\begin{itemize}
\item Digit: \{“The image shows a [digit]”, “The digit appears to be [digit]”,  “There is an image showing a [digit]”, “The number is a [digit]”\}
\item Digit Color: \{“The color is [color]”, “The digit appears to be [color]”, “There is a [color] image”, “The image is [color]”\}
\item Shape: \{“The shape is a [shape]”, “The shape appears to be a [shape]”, “There is a [shape]”, “The image has a [shape]”\}
\item Shape Size: \{“The shape size is [size]”, “The size of the shape is [size]”, “The shape is [size]”\}
\end{itemize}
In order to construct the final description, the generated sentences are shuffled and any duplicate sentences are pruned.

\subsection{Extended Evaluations}

\begin{figure*}
\begin{center}
\includegraphics[width=0.93\textwidth]{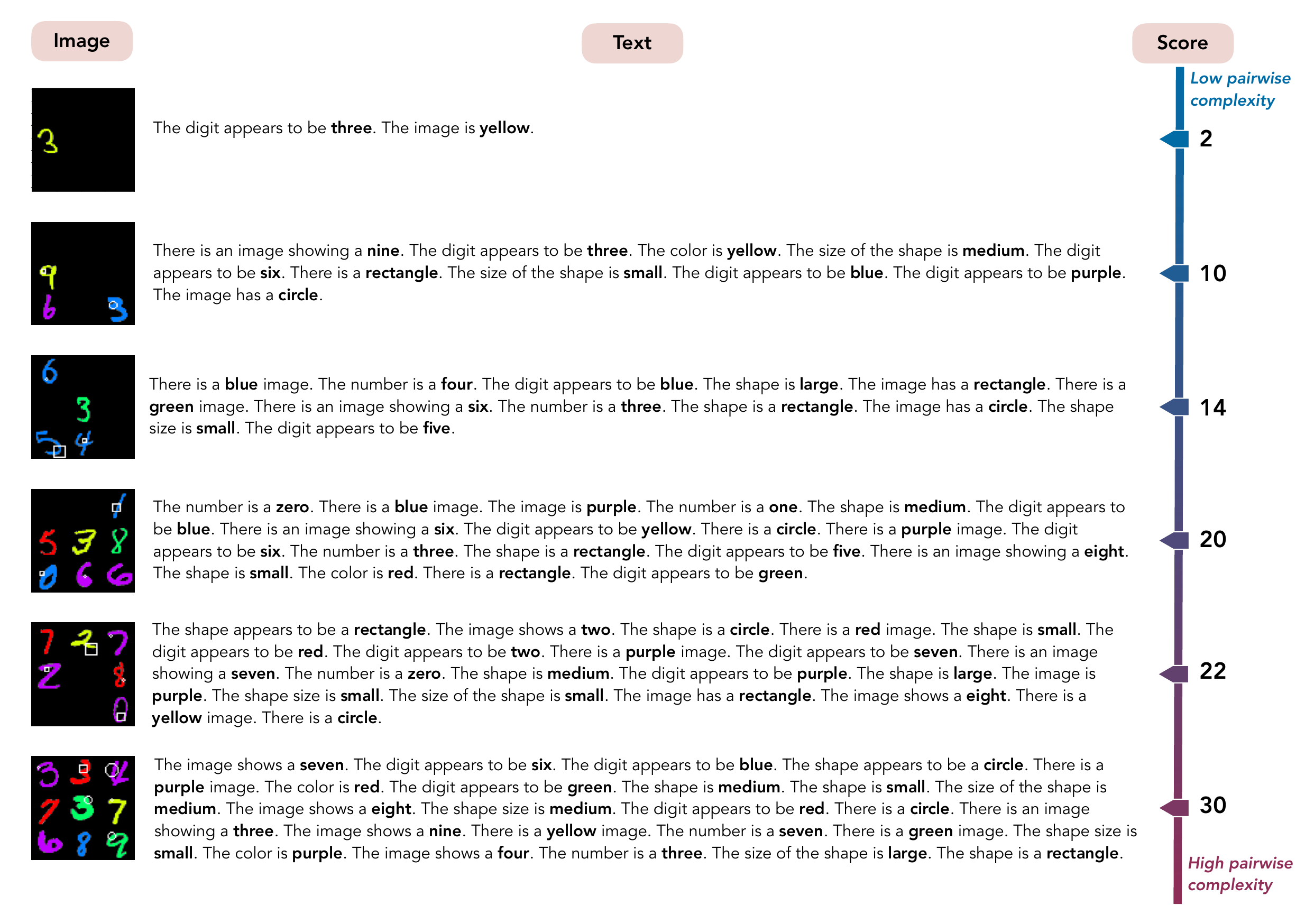}
\end{center}
\caption{Examples of randomly-selected image-text pairs and the associated pairwise complexity scores from each of the six \textsc{DocMNIST} training datasets. Each textual attribute (bolded) appears in at least one region of the image. The image-text pairwise complexity score measures the number of distinct region-attribute pairs.}
\label{fig:appendixdocmnist}
\end{figure*}

\noindent
In order to systematically evaluate the role of training dataset complexity on the fine-grained reasoning ability of a standard one-to-one VLM, we generate a set of six \textsc{DocMNIST} training datasets that vary in average image-text pairwise complexity. In Figure \ref{fig:appendixdocmnist}, we provide examples of randomly-selected image-text pairs from each of the six training datasets. 

\smallskip
\smallskip
\noindent
For each generated \textsc{DocMNIST} dataset, we train a standard one-to-one VLM to contrastively learn alignments between images and the associated text. The image encoder consists of a ResNet-50 model initialized with pretrained CLIP RN50 weights \cite{radford-clip, resnet2015}. The output of the encoder is an $L_2$-normalized embedding of dimension 1024.  The text encoder consists of a pretrained CLIP text encoder with frozen weights. Since textual descriptions may be longer than the maximum token limit of the text encoder, we split each description into sentences, compute the embedding of each sentence, and then average the embeddings together; this yields a single $L_2$-normalized text embedding of dimension 1024. The VLM is optimized using a standard bidirectional contrastive loss function with a temperature of 0.07. We train on a single NVIDIA V100 GPU for 100 epochs with early stopping if the loss fails to decrease for 5 consecutive epochs. We use an initial learning rate of 5e-5 and a batch size of 256. We repeat this procedure with three different random seeds for each generated \textsc{DocMNIST} dataset.  

\smallskip
\smallskip
\noindent
At inference time, we construct a fixed \textsc{DocMNIST} test set with 1196 images; the images consist of a total of 5982 regions. We evaluate our trained VLMs on both a fine-grained text $\rightarrow$ region retrieval task and a region $\rightarrow$ text retrieval task. We emphasize here that although the VLMs were trained at the \textit{image-level}, we are specifically evaluating the ability of the models to understand \textit{region-level} features. 

\smallskip
\smallskip
\noindent
Given a textual query, the text $\rightarrow$ region retrieval task involves retrieving image regions that capture the content of the query. To this end, we use the trained VLM to generate an embedding for each individual region in the \textsc{DocMNIST} test set. Then, for each of the twenty attributes in set $A$, we generate query sentences using the templates defined in the previous section; this yields a total of twenty textual query embeddings. We use dot-product similarity to identify the top-$k$ regions from the DocMNIST test set that are most similar to each query. In Figure \ref{fig:docmnistresults}, we report the R-Precision metric, which defines $k$ as the total number of ground-truth regions in the \textsc{DocMNIST} test set that possess the queried attribute. 

\smallskip
\smallskip
\noindent
We follow a similar procedure for the region $\rightarrow$ text retrieval task. For each of the 5.9k regions in the \textsc{DocMNIST} test set, the region $\rightarrow$ text retrieval task determines if we can identify the textual attributes depicted in the region. We compute the dot-product similarity between the region embedding and the 20 attributes in set $A$. We note that a given region includes at most one attribute from each of the four categories (as defined in Section \ref{appendixsec:docmnistimp}); as a result, we identify the top-scoring attribute from each of the four categories (digit, digit color, shape, and shape size). In Figure \ref{fig:docmnistresults}, we report the R-precision metric, which determines the percentage of the top-$k$ identified attributes that are accurate; here, $k$ is defined as the total number of ground-truth attributes in the region, which will range between 2 and 4.  

\begin{figure*}
\begin{center}
\includegraphics[width=0.93\textwidth]{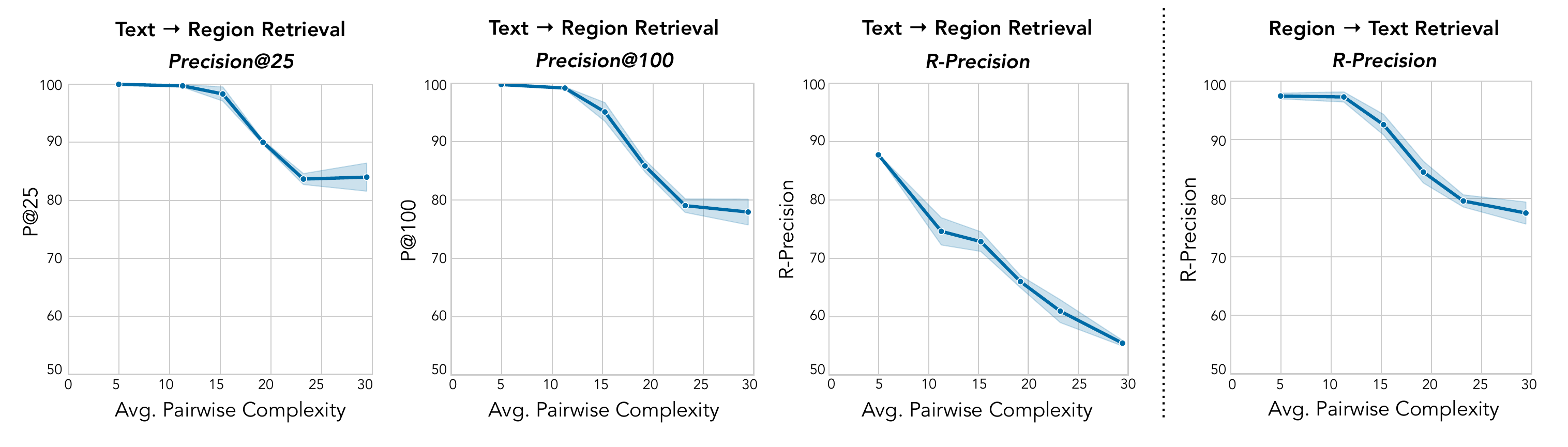}
\end{center}
\caption{We expand on the results shown in Figure \ref{fig:docmnistresults} by providing Precision@25 and Precision@100 metrics for the text $\rightarrow$ region retrieval task. Across all metrics, representations generated using one-to-one VLMs exhibit significant performance degradations as the average pairwise complexity of the dataset increases.}
\label{fig:appendixdocmnistgraphs}
\end{figure*}

\smallskip
\smallskip
\noindent
In Figure \ref{fig:appendixdocmnistgraphs}, we expand on the results shown in Figure \ref{fig:docmnistresults} by providing Precision@25 and Precision@100 metrics for the text $\rightarrow$ region retrieval task. Additional details on the fine-grained text $\rightarrow$ region and region $\rightarrow$ text retrieval tasks are provided in Appendix Section \ref{appendixsec:eval}. 

\section{\name{} Implementation Details}
\label{appendix:villaimplementation}
\noindent
In this section, we provide additional implementation details for \name{}. 
\subsection{Datasets}
\label{appendixsec:datasetdetails}

\noindent
In this work, we use four training datasets across various real-world domains: DocMNIST (synthetic images), DeepFashion (product data), MIMIC-CXR (medical images), and COCO (natural images). Additional details on each dataset are provided below:

\smallskip
\smallskip
\noindent
\textbf{DocMNIST}: We create the synthetic \textsc{DocMNIST} dataset using the procedure described in Section \ref{sec:understanding}. In Section \ref{sec:understanding}, we generated six versions of the \textsc{DocMNIST} training set with varying complexities; for the remainder of this work, we use the version of the \textsc{DocMNIST} training set with the highest complexity. Specifically, the training set has an average pairwise complexity of $c=29.4$ with a total of $b=300K$ attributes. We also generate a validation dataset with $b=10k$ attributes. 

\smallskip
\smallskip
\noindent
\textbf{DeepFashion} \cite{deepfashiondata2016,jiang2022text2human}: The DeepFashion-Multimodal dataset consists of 44,096 high-resolution human images obtained from clothing retail websites. We filter the dataset to include 42,537 images that have associated textual descriptions, and we resize all images to $3 \times 335 \times 228$. We randomly assign 70\% of the dataset (corresponding to 29,694 images) to the training set, 10\% of the dataset (corresponding to 3985 images) to the validation set, and the remaining 20\% of the dataset (corresponding to 8858 images) to the test set. Each image is annotated with structured labels, which we filter to include 58 labels divided across 16 categories as follows: glasses (eyeglasses, sunglasses, glasses in hand or on clothes), hat (yes), lower clothing length (long, medium short, three-quarter, three-point), lower color (lattice, pure color, floral, color block, graphic, striped), lower fabric (cotton, chiffon, leather, knitted, denim), neckline (square, lapel, round, standing, v-neck, suspenders), neckwear (yes), outer color (pure color, graphic, color block, floral, lattice, striped), outer fabric (cotton, chiffon, knitted, denim, leather), ring (yes), sleeve length (sleeveless, medium-sleeve, short-sleeve, long-sleeve), socks (leggings, socks), upper color (floral, pure color, graphic, lattice, color block, striped), upper fabric (leather, chiffon, knitted, cotton, denim), waist accessory (belt, some accessory), wrist accessory (yes). 

\smallskip
\smallskip
\noindent
\textbf{MIMIC-CXR} \cite{johnson2019mimic, physionet}: The MIMIC-CXR dataset consists of 377,110 chest X-ray images and associated physician reports obtained from the Beth Israel Deaconess Medical Center. Following prior work \cite{huang2021gloria}, we filter the dataset to include only frontal views (AP or PA). We also remove any images from the dataset that do not have a valid report; here, we define a valid report as one that has a non-empty Findings or Impression section. The final dataset includes a total of 231,564 images, and we resize all images to $3 \times 256 \times 256$. We randomly assign 70\% of the dataset (162,417 images) to the training set, 10\% of the dataset (23,592 images) to the validation set, and 20\% of the dataset (45,555 images) to the test set. Since medical reports often include a large sentences with negative findings (\eg \textit{the patient does not have pneumonia}), we use RadGraph to remove any sentences discussing absent entities \cite{radgraph}. The final textual description includes all remaining sentences from the Findings and Impression sections.

\smallskip
\smallskip
\noindent
\textbf{COCO} \cite{coco2014}: The Microsoft COCO training dataset consists of 114,648 natural images, each associated with five captions. We use the train2017 data split for training and the val2017 data split for inference. 

\smallskip
\smallskip
\noindent
In order to estimate the average image-text pairwise complexity of each dataset (as shown in Table \ref{tab:dataset}), we compute the number of ground-truth region-attribute pairs in each image as follows. 
\begin{itemize}
\item \textit{DocMNIST}: As detailed in Section \ref{sec:understanding}, the average image-text pairwise complexity of the \textsc{DocMNIST} dataset is 29.4. 
\item \textit{DeepFashion}: The DeepFashion-Multimodal dataset does not provide ground-truth region-attribute annotations. As a result, we provide a rough estimate of image-text pairwise complexity; it is important to note that this value may fluctuate depending on the selection of attributes and regions. We use the provided structured labels in the dataset as the attribute set, and we determine appropriate locations for each label from the following set of regions: head, upper body, hands, lower body, and feet. We count the number of attribute-region paris in each image and determine the average pairwise complexity to be 7.9. 
\item \textit{MIMIC-CXR}: Similar to DeepFashion, the MIMIC-CXR dataset does not provide ground-truth region-attribute annotations. As a result, we provide a rough estimate of pairwise complexity; again, it is important to note that this value may fluctuate depending on the selection of attributes and regions. In order to select attributes, we use RadGraph \cite{radgraph} to extract all entities from the reports (filtered to only include nouns). We then use a series of hand-crafted rules to parse sentences for location-related information, and we determine associations between identified entities and nine regions: left upper lung, left lower lung, left lung, right upper lung, right lower lung, right lung, heart, osseous structures, and stomach. We count the number of attribute-region pairs in each image and determine the average image-text pairwise complexity to be 5.0. 
\item \textit{COCO}: We estimate pairwise complexity by using the ground-truth region annotations provided in the COCO training set. We count the number of ground truth region-attribute pairs in each image and determine the average image-text pairwise complexity of the COCO dataset to be 6.8. 
\end{itemize}

\begin{figure*}
\begin{center}
\includegraphics[width=0.93\textwidth]{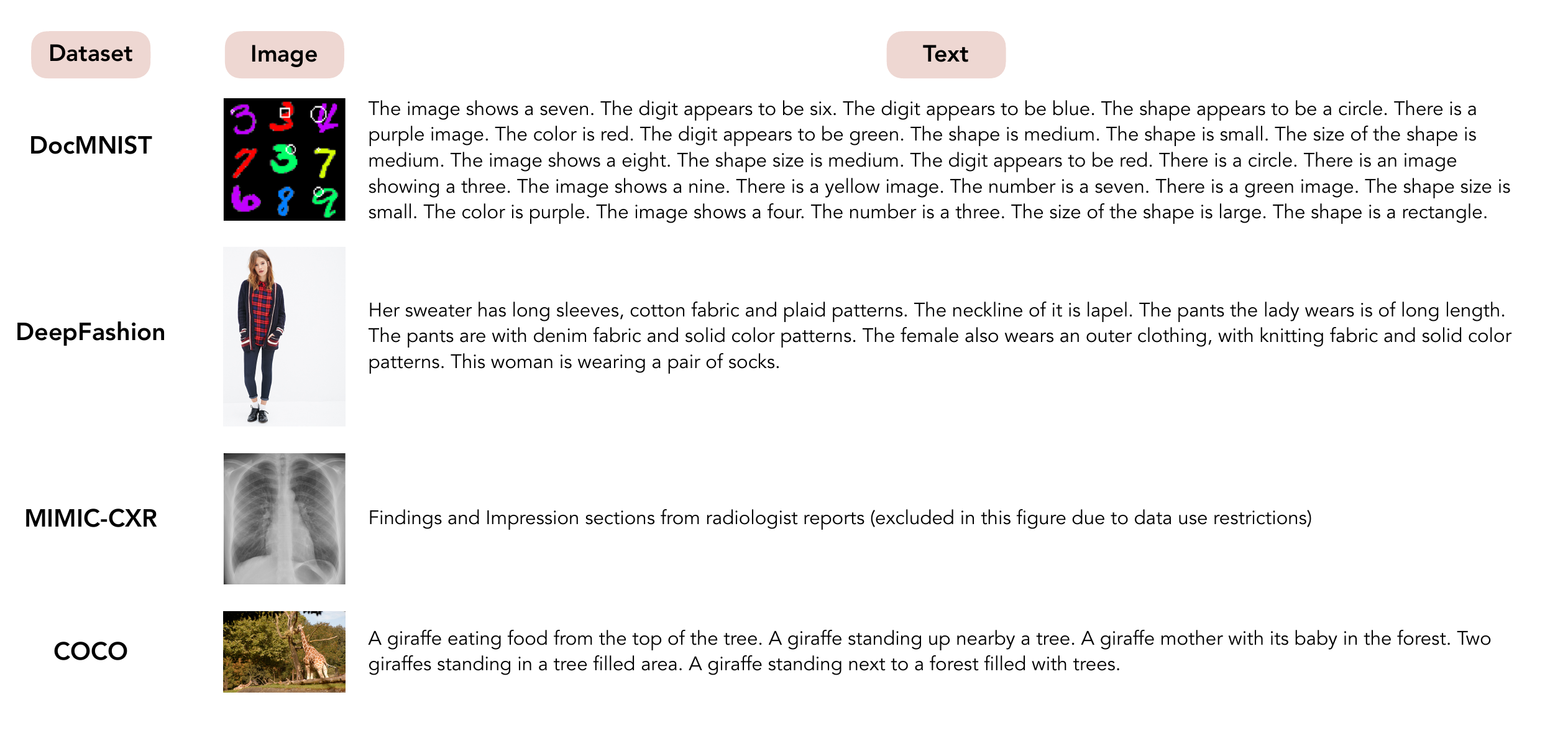}
\end{center}
\caption{Examples of randomly-selected image-text pairs from each of the four training datasets: DocMNIST, DeepFashion, MIMIC-CXR, and COCO. Note that due to data use restrictions for the MIMIC-CXR dataset, we provide a representative example of a chext X-ray from the web and do not provide a text sample.}
\label{fig:datasetexamples}
\end{figure*}

\subsection{Extended Description of the Mapping Model (Stage 1)}
\label{appendixsec:mappingmodeldesc}

\noindent
\textbf{Decomposing Images and Text:} We decompose each image $x_i$ into $r_i$ regions, expressed as $x_i = \{x_i^0, x_i^1,…,x_i^{r_i}\}$. There are a variety of ways in which an image can be decomposed into regions, such as dividing images into equal-sized segments (\eg quadrants) or using region proposal networks (RPNs). Note that we place no restrictions on the size or coverage of regions; regions can be of any size and can overlap. Ideally, regions should capture the key features in the image. We explore a variety of region selection methods across our four datasets, which we discuss in further detail below:
\begin{itemize}
\item \textit{DocMNIST:} As described in Section \ref{sec:understanding}, each image is composed of 9 equally-sized candidate regions of size $3 \times 28 \times 28$. We only consider regions with at least one assigned attribute.
\item \textit{DeepFashion:} Since the DeepFashion dataset consists of human images, we divide each image lengthwise into 4 equally-sized regions. Roughly, these regions correspond to the head, torso, legs, and feet; however, some variation is expected due to poses. 
\item \textit{MIMIC-CXR:} We train a custom region proposal network to divide each image into 3 anatomic regions: right lung, left lung, and heart. To train the heart segmentation network, we used the JSRT dataset (\url{http://db.jsrt.or.jp/eng.php}) which contains 247 chest x-rays with heart segmentation masks. To train the lung segmentation network, we used the JSRT dataset (\url{http://db.jsrt.or.jp/eng.php}), which contains 247 chest x-rays with lung segmentation masks, and two additional datasets published by the U.S. National Library of Medicine \cite{jaeger2014chest,Candemir2014}, which contain a total of 566 chest x-rays with lung segmentation masks. Each image in these datasets was preprocessed with the following operations, performed in sequence: normalization to the range [0,1]; resized to size (224, 224); histogram equalization. Both the lung and heart segmentation networks were trained with a batch size of 16 for 150 epochs using a UNet \cite{unet} architecture and a learning rate of 1e-4. We used random brightness contrast, gaussian blur, and affine transforms as augmentations during training. The predicted segmentation masks were postprocessed with the following operations, performed in sequence: binary opening operation with a disk (radius=5) structuring element; keeping only the largest contiguous predicted segment; binary fill holes; binary dilation with a disk (radius=5) structuring element.
\item \textit{COCO:} We use a pretrained RPN with identical settings to prior work \cite{Zhong_2022_CVPR_regionclip}. The RPN generates 300 candidate regions for each image. For training the mapping model (stage 1), we select 20 regions from the set of 300 such that selected regions share minimal overlap. For training the VLM (stage 2), we randomly sample 100 regions from the set of 300. 
\end{itemize}

\smallskip
\smallskip
\noindent
Similarly, we decompose each textual description $t_i$ into $a_i$ attributes, expressed as $t_i = \{t_i^0, t_i^1, …, t_i^{a_i}\}$. Dataset-specific implementation details are provided below:
\begin{itemize}
    \item \textit{DocMNIST}: As described in Section \ref{sec:understanding}, the \textsc{DocMNIST} dataset includes a total of 20 attributes: zero, one, two, three, four, five, six, seven, eight, nine, purple, blue, green, yellow, red, rectangle, circle, small, medium, and large. We extract attributes in this set from the textual description. 
    \item \textit{DeepFashion:} We use the 58 provided structured labels (described in Section \ref{appendixsec:datasetdetails}) as our set of relevant attributes. 
    \item \textit{MIMIC-CXR:} We use RadGraph, an off-the-shelf entity extractor, to identify entities in each textual description \cite{radgraph}. RadGraph classifies each entity with one of three labels: Definitely Present, Uncertain, and Definitely Absent; we filter the set of identified entities to only include those that are Definitely Present. We then use the spaCy library to filter the set of entities to those that are nouns. Finally, we identify the 50 entities that occur most frequently in the training dataset as our final set of attributes. 
    \item \textit{COCO:} In line with with prior work, we extract 4.7k textual attributes (e.g. giraffe, man, bicycle, etc.) from the captions \cite{Zhong_2022_CVPR_regionclip}. 
\end{itemize}

\smallskip
\smallskip
\noindent
\textbf{Representing Regions and Attributes:} We generate embeddings for image regions using the procedure described in Section \ref{subsec:mapping}. As described in Section \ref{subsec:mapping}, the output of the image encoder is passed through a series of $p$ projection heads, where each projection head consists of a linear layer, a ReLU function, and a second linear layer. For our experiments with DocMNIST, DeepFashion, and MIMIC-CXR, we set $p$ to be equal to the total number of attributes in the dataset; as a result $p$ equals 20, 58, and 50 respectively. We find that using a distinct projection heads for each attribute allows the model to learn fine-grained differences between similar attributes (\eg distinguishing cotton and chiffon fabric in DeepFashion). This is particularly useful for real-world datasets, where images often exhibit high inter-class similarity \cite{zhang2020convirt}. However, using distinct projection heads for each attribute may be ineffective if (a) there are a large number of attributes in the dataset or (b) there is high diversity across images and attributes; in these cases, we find that models can effectively learn patterns with a fewer number of projection heads. Consequently, for our experiments with COCO, we use a single projection head ($p = 1$). 

\smallskip
\smallskip
\noindent
Additionally, since the COCO dataset is comprised of natural images and likely exhibits a similar distribution to the CLIP pretraining dataset, we find that an off-the-shelf CLIP model (with no additional tuning) can map regions to attributes with relatively high accuracy (as shown in Table \ref{tab:mapping_acc}). As a result, in order to capture the knowledge of the CLIP model, we compute the final embedding as a weighted average of (a) the input embedding provided to the projection head and (b) the output embedding generated by the projection head; this approach is similar to prior work on feature adapters \cite{adapter}. Our approach generates more accurate region-attribute mappings than the original off-the-shelf CLIP model (by 4.9 points as shown in Table \ref{tab:mapping_acc}). 

\smallskip
\smallskip
\noindent
We generate embeddings for attributes using the procedure described in Section \ref{subsec:mapping}. We insert each attribute into pre-defined prompt templates and extract representations from a pretrained text encoder. Below, we provide dataset-specific implementation details:
\begin{itemize}
\item \textit{DocMNIST:} We insert each attribute into the prompt templates defined in Section \ref{appendixsec:docmnistimp}. We represent each attribute as the average of its associated prompt embeddings. 
\item \textit{DeepFashion:} We insert each attribute into the following pre-defined prompt templates, where components in square brackets are replaced with the attribute: \{“The person wears a [attribute]”, “There is a [attribute]”, “The person is wearing [attribute]”, “The man is wearing a [attribute]”, “The woman is wearing a [attribute]”, “The clothing is [attribute]”\}. We represent each attribute as the average of its associated prompt embeddings. 
\item \textit{MIMIC-CXR:} For each attribute, we identify all sentences in the training dataset that mention the attribute and then filter the list to include the 200 most frequently-occurring sentences. We represent each attribute as the average of its associated sentence embeddings.
\item \textit{COCO:} We insert each attribute into 81 prompt templates defined in a prior study \cite{Zhong_2022_CVPR_regionclip}. Sample prompt templates include: \{“A photo of a [attribute]”, “A good photo of the [attribute]”, “A photo of the small [attribute]”\}. We represent each attribute as the average of its associated prompt embeddings.
\end{itemize}

\smallskip
\smallskip
\noindent
\textbf{Training Procedure:} The mapping model is optimized with the following batch-wise loss function (where $L(x_i, t_i)$ is defined in Section \ref{subsec:mapping}).
$$L_B = \frac{1}{\sum_{j=1}^{|B|} |t_j|} \sum_{i=1}^{|B|} L(x_i, t_i)$$

\smallskip
\smallskip
\noindent
We train the mapping model on a single NVIDIA A100 GPU with an initial learning rate of 1e-4. Dataset-specific training details are provided below, with hyperparameters selected based on mapping accuracy on a validation set as well as GPU memory constraints: 
\begin{itemize}
\item \textit{DocMNIST}: We use a batch size of 48 and train for 30 epochs. We set the loss temperature as $\tau=0.07$. 
\item \textit{DeepFashion}: We use a batch size of 48 and train for 20 epochs. We set the loss temperature as $\tau=0.07$. 
\item \textit{MIMIC-CXR}: We use a batch size of 48 and train for 5 epochs. We set the loss temperature as $\tau=1$. 
\item \textit{COCO}: We use a batch size of 24 and train for 5 epochs. We set the loss temperature as $\tau=0.07$. 
\end{itemize}

\subsection{Extended Description of the VLM Model (Stage 2)}
\label{appendixsec:stage2details}
\noindent
We use the mapping model from Section \ref{subsec:mapping} to assign attributes to regions as follows. For DocMNIST, DeepFashion, and MIMIC-CXR, our approaches for selecting regions yield regions with zero, one, or multiple attributes. In order to account for a variable number of attributes that may be associated with each region, we use the following procedure to map attributes to regions. For a sample $(x_i, t_i)$ and textual attribute $k \in t_i$, we compute the pairwise dot product between $P_k(e_i)$ and $h_i^k$, resulting in a score vector $v \in \mathbb{R}^{r_i \times 1}$. We then assign $k$ to all regions with a score greater than $max(v) - \epsilon$, where $\epsilon$ is a pre-defined threshold. As desired, this procedure assigns zero, one, or multiple attributes to a region. We assign the value of $\epsilon$ by evaluating the quality of the region-attribute mappings on a validation set (details on this evaluation are provided in Appendix Section \ref{appenddixsec:evaluatingregatt}). We select $\epsilon=0.2$ for DocMNIST, $\epsilon=0.2$ for DeepFashion, and $\epsilon=0.1$ for MIMIC-CXR. 

\smallskip
\smallskip
\noindent
For COCO, we select candidate regions using an RPN, which generally yields tight bounding boxes that capture a single attribute.  In this case, we invert the above procedure and instead assign regions to attributes. For a sample $(x_i, t_i)$ and textual attribute $k \in t_i$, we compute the pairwise dot product between $P_k(e_i)$ and $h_i^k$. We repeat this computation for all textual attributes in $t_i$, resulting in a final score vector $v \in \mathbb{R}^{r_i \times a_i}$. We then assign each region to its highest-scoring attribute. As desired, this procedure assigns each region to a single attribute. Additionally, using this procedure ensures that our work is consistent with prior object detection models trained on the COCO dataset \cite{Zhong_2022_CVPR_regionclip}. 

\smallskip
\smallskip
\noindent
For each generated region-attribute pair, we replace the attribute with the segment of the original textual description containing the attribute; for example, the attribute “red” may be replaced with the sentence “The digit appears to be red” from the original description. We perform this step in order to allow for better reasoning over textual cues when training the VLM. 

\smallskip
\smallskip
\noindent
We then augment the training dataset to include generated region-attribute pairs in addition to the original image-text samples. We use the augmented dataset to train a one-to-one VLM. Below, we provide implementation details for each dataset. 
\begin{itemize}
\item \textit{DocMNIST}: The image encoder is initialized with weights from a pretrained CLIP ResNet-50. The text encoder takes the form of a pretrained CLIP text encoder with frozen weights. We optimize the VLM using a standard bidirectional contrastive loss function with the temperature parameter set as $\tau=0.07$. We train the VLM on a single NVIDIA V100 GPU with an initial learning rate of 5e-5 and a batch size of 256. 
\item \textit{DeepFashion}: The image encoder is initialized with weights from a pretrained CLIP ResNet-50. The text encoder takes the form of a pretrained CLIP text encoder with frozen weights. We optimize the VLM using a standard bidirectional contrastive loss function with the temperature parameter set as $\tau=0.07$. We train the VLM on a single NVIDIA V100 GPU with an initial learning rate of 5e-5 and a batch size of 96. 
\item \textit{MIMIC-CXR}: The image encoder is initialized with weights from a pretrained ConVIRT ResNet-50 using the implementation provided in ViLMedic \cite{vilmedic2022}. The text encoder takes the form of a pretrained SBERT text encoder with frozen weights. We optimize the VLM using a standard bidirectional contrastive loss function with the temperature parameter set as $\tau=1.0$. We train the VLM on a single NVIDIA V100 GPU with an initial learning rate of 5e-5 and a batch size of 196. 
\item \textit{COCO}: We use the training framework and pretrained configurations provided by \cite{Zhong_2022_CVPR_regionclip}. The image encoder is initialized with weights from a pretrained CLIP ResNet-50. The text encoder takes the form of a pretrained CLIP text encoder with frozen weights. The VLM is optimized using both a contrastive loss function and a distillation loss function, as implemented by \cite{Zhong_2022_CVPR_regionclip}. We modify the contrastive loss function provided by \cite{Zhong_2022_CVPR_regionclip} as follows: given a region-attribute pair, we expand the negative set to include \textit{all} attributes, rather than just the other attributes in the batch. We train the VLM on 8 NVIDIA A100 GPUs with an initial learning rate of 0.002 and a batch size of 96. 
\end{itemize}

\section{Experimental Details}
\label{appendixsec:eval}
\noindent
In this section, we provide additional details on our experiments.

\subsection{Evaluation Tasks}
\noindent
Here, we provide further details on our three evaluation tasks: zero-shot object detection, text $\rightarrow$ region retrieval, and region $\rightarrow$ text retrieval.
\subsubsection{Zero-Shot Object Detection}
\noindent
We compare \name{} with three prior zero-shot object detection methods; we provide additional details on these methods below. 
\begin{itemize}
\item \textit{OVR-CNN} \cite{zareian2021_ovr}: OVR-CNN is an object detector trained with both annotated region-category pairs as well as image-caption pairs. OVR-CNN is trained using 114k image-text pairs from the COCO dataset.
\item \textit{CLIP} \cite{radford-clip}: CLIP is a VLM that leverages contrastive self-supervised learning to learn relationships between paired image-text samples. We use a CLIP model with a ResNet-50 backbone, which was trained on 400 million image-text pairs. 
\item \textit{RegionCLIP} \cite{Zhong_2022_CVPR_regionclip}: RegionCLIP was recently introduced for improving the fine-grained reasoning ability of CLIP. RegionCLIP uses the pretrained CLIP model in a zero-shot fashion to match candidate image regions with plausible textual attributes; these mappings are then used during training. We compare against two versions of RegionCLIP: (a) a version trained on 3 million image-text pairs from the Conceptual Captions (CC3M) dataset and (b) a version trained on 114k image-text pairs from the COCO dataset. Our approach is most comparable to version (b); however, this model is not made publicly available, resulting in some missing values in the corresponding row in Table \ref{tab:obj}.  
\end{itemize}

\subsubsection{Text $\rightarrow$ Region Retrieval}
\label{appendixsec:textregionretrieval}
\noindent
The text$\rightarrow$region retrieval task evaluates the ability of a VLM to reason over fine-grained relationships between image regions and textual attributes. Given a textual query (\eg “The person is wearing a hat”), the text$\rightarrow$region retrieval task determines if we can retrieve image regions that capture the content of the query. Below, we provide implementation details for the text$\rightarrow$region retrieval task on the \textsc{DocMNIST} and DeepFashion datasets:
\begin{itemize}
\item \textit{DocMNIST}: We consider 20 textual queries for DocMNIST, with each query representing a distinct attribute from the following set: \{zero, one, two, three, four, five, six, seven, eight, nine, purple, blue, green, yellow, red, rectangle, circle, small, medium, large\}. We generate textual queries by inserting each attribute into the prompt templates provided in Appendix Section \ref{appendixsec:docmnistimp}. 
\item \textit{DeepFashion}: We construct the text$\rightarrow$region retrieval task by extracting 7.9k regions from the DeepFashion test set. The DeepFashion dataset includes human-parsing labels for a subset of the images, where various features of each image (\eg glasses, hat, etc.) are labeled with pixel-level annotations in the image; we use the human-parsing labels to identify the ground-truth attributes in each region. We then construct 46 textual queries for DeepFashion, with each query representing a distinct attribute from the set listed in Section \ref{appendixsec:datasetdetails}. Although the original attribute set includes 58 attributes across 16 categories, we include only (a) attributes that occur at least once in the retrieval set of 7.9k regions and (b) attributes that are included in the provided human-parsing labels; this results in a total of 46 attributes across 14 categories. For the Precision@25 and Precision@100 metrics reported in Table \ref{tab:ret}, we further filter this list to only include attributes that occur at least 25 and 100 times in the retrieval set of 7.9k regions; this yields 40 and 25 attributes respectively.  For each attribute, we construct a textual query by inserting the attribute into a custom prompt template. Prompt templates were designed to emulate linguistic patterns in the training set and are constructed based on attribute categories, as shown below (categories are italicized and attributes are in parentheses).   
\begin{itemize}
\item \textit{glasses} (sunglasses): “The person wears sunglasses”
\item \textit{hat} (yes): “The person wears a hat”
\item \textit{lower color} (floral, pure color, graphic, lattice, color block, striped): “The lower clothing has [attribute] patterns”
\item \textit{lower fabric} (leather, chiffon, knitted, cotton, denim): “The lower clothing is [attribute]”
\item \textit{neckline} (square, lapel, round, standing, v-neck, suspenders): “The neckline is [attribute]”
\item \textit{neckwear} (yes): “The person has neckwear”
\item \textit{outer color} (floral, pure color, graphic, lattice, color block, striped): “The sweater has [attribute] patterns”
\item \textit{outer fabric} (leather, chiffon, knitted, cotton, denim): “The outer clothing is [attribute]”
\item \textit{ring} (yes): “The person is wearing a ring”
\item \textit{socks} (leggings, socks): “The person wears [attribute]”
\item \textit{upper color} (floral, pure color, graphic, lattice, color block, striped): “The upper clothing has [attribute] patterns”
\item \textit{upper fabric} (chiffon, knitted, cotton, denim): “The upper clothing is [attribute]”
\item \textit{waist accessory} (belt): “The person is wearing a belt”
\item \textit{wrist accessory} (yes): “There is an accessory on the wrist”
\end{itemize}

\end{itemize}

\noindent
We compare \name{} with four baselines; we provide additional details on these methods below. 
\begin{itemize}
\item \textit{CLIP-ZS} \cite{radford-clip}: The CLIP-ZeroShot (CLIP-ZS) baseline applies an off-the-shelf CLIP model to the text$\rightarrow$region retrieval task in a zero-shot manner. We use a CLIP model with a ResNet-50 visual backbone. We encode each textual query using the CLIP text encoder; then, the regions with the highest dot-product similarities are identified. 
\item \textit{CLIP-FT-Img}: The CLIP-FineTuned-Image (CLIP-FT-Img) baseline applies a fine-tuned CLIP model to the text$\rightarrow$region retrieval task. We fine-tune a CLIP model with a ResNet-50 visual backbone on image-text pairs from the DocMNIST and DeepFashion training datasets. The model is optimized using a standard bidirectional contrastive loss function with a temperature of 0.07. We train on a single NVIDIA V100 GPU for 100 epochs with early stopping if the loss fails to decrease for 5 consecutive epochs. We use an initial learning rate of 5e-5 and a batch size of 256 for DocMNIST and 96 for DeepFashion. In order to perform text$\rightarrow$region retrieval, we encode each textual query using the CLIP text encoder and identify regions with the highest dot-product similarities. 
\item \textit{CLIP-FT-Reg}: The CLIP-FineTuned-Region (CLIP-FT-Reg) baseline applies a fine-tuned CLIP model to the text$\rightarrow$region retrieval task. We fine-tune a CLIP model with a ResNet-50 visual backbone on both image-text and region-text pairs, where each region is aligned with the entire textual description. The model is optimized using a modified bidirectional contrastive loss function with a temperature of 0.07. Since each batch now includes multiple regions that share the same textual description, we modify the contrastive loss function such that for a given region or textual description, the negative set will not include any other regions or textual descriptions from the same image. We train on a single NVIDIA V100 GPU for 100 epochs with early stopping if the loss fails to decrease for 5 consecutive epochs. We use an initial learning rate of 5e-5 and a batch size of 256 for DocMNIST and 96 for DeepFashion. In order to perform text$\rightarrow$region retrieval, we encode each textual query using the CLIP text encoder and identify regions with the highest dot-product similarities.  
\item \textit{CLIP-ZS-Map} \cite{radford-clip,Zhong_2022_CVPR_regionclip}: The CLIP-ZeroShot-Mapping (CLIP-ZS-Map) baseline first applies CLIP in a zero-shot manner to generate region-attribute pairs (following the procedure in Section \ref{subsec:train}). We set $\epsilon=0.1$ for DocMNIST and $\epsilon=0.01$ for DeepFashion (details on the selection $\epsilon$ are provided in Appendix Sections \ref{appendixsec:stage2details} and \ref{appenddixsec:evaluatingregatt}). Then, a CLIP model with a ResNet-50 visual backbone is fine-tuned with image-text pairs as well as the generated region-attribute pairs. This baseline is comparable to RegionCLIP \cite{Zhong_2022_CVPR_regionclip}. The model is optimized using a standard bidirectional contrastive loss function with a temperature of 0.07. We train on a single NVIDIA V100 GPU for 100 epochs with early stopping if the loss fails to decrease for 5 consecutive epochs. We use an initial learning rate of 5e-5 and a batch size of 256 for DocMNIST and 96 for DeepFashion. In order to perform text$\rightarrow$region retrieval, we encode each textual query using the CLIP text encoder and identify regions with the highest dot-product similarities.  
\end{itemize}

\subsubsection{Region $\rightarrow$ Text Retrieval}
Given an image region, the region$\rightarrow$text retrieval task determines if we can retrieve the textual attributes depicted in the region. We follow a similar set-up to the text$\rightarrow$region retrieval task detailed in Appendix Section \ref{appendixsec:textregionretrieval}, and we compare \name{} with the same baselines: CLIP-ZS, CLIP-FT-Img, CLIP-FT-Reg, and CLIP-ZS-Map. We note that for the DocMNIST dataset, a given region includes at most one attribute from each of the four categories (as defined in Section \ref{appendixsec:docmnistimp}); as a result, we retrieve at most one attribute from each of the four categories (digit, digit color, shape, and shape size). In Figure \ref{fig:docmnistresults}, we report the R-precision metric, which determines the percentage of the top-$k$ retrieved attributes that are accurate; here, $k$ is defined as the total number of ground-truth attributes in the region.

\smallskip
\smallskip
\noindent
We additionally evaluate retrieval performance on the CheXpert 5x200 benchmark, which consists of 1000 chest X-rays across five disease categories (atelectasis, cardiomegaly, consolidation, pulmonary edema, and pleural effusion) \cite{huang2021gloria}. The dataset is class-balanced, with 200 chest X-rays corresponding to each disease; additionally, the chest X-rays were selected such that no X-ray depicts more than one disease. Disease labels are converted into textual phrases using pre-defined prompts. Then, given a chest X-ray, the goal is to retrieve the textual phrase corresponding to the correct disease. In Table \ref{tab:chexpert}, we report retrieval accuracy (which is equivalent to a Precision@1 score) in line with prior work \cite{huang2021gloria}. 

\smallskip
\smallskip
\noindent
In order to convert disease labels into text, we use the same process that we use to generate attribute embeddings for MIMIC-CXR. For each disease label, we use the filtered set of RadGraph entities created in Section \ref{appendixsec:mappingmodeldesc} to identify all sentences in the training dataset that mention the disease; we then filter the list to include the 200 most frequently-occurring sentences. We represent each disease as the average of its associated sentence embeddings. If a disease label does not occur in the filtered RadGraph entity set, we instead represent the disease by encoding the following simple prompt: “patient with [disease]”. 

\smallskip
\smallskip
\noindent
Although the CheXpert 5x200 task has been previously considered as an image$\rightarrow$text retrieval task, we instead formulate it as a region$\rightarrow$text retrieval task when evaluating \name{}. We do so by considering each X-ray as a set of four regions - right lung, left lung, heart, full image - and using \name{} to generate embeddings for each region; then, we perform retrieval by computing the maximum pairwise similarity with the text phrases. Note that for our baselines, we follow the standard image$\rightarrow$text retrieval approach in order to remain consistent with prior work \cite{huang2021gloria}. 

\smallskip
\smallskip
\noindent
We compare \name{} with three baselines; we provide additional details on these methods below. Of these baselines, we note that ConVIRT and BioViL can be classified as one-to-one VLMs; however, GLoRIA is not a one-to-one VLM due to the use of a fine-grained local contrastive loss function. 
\begin{itemize}
\item \textit{ConVIRT} \cite{zhang2020convirt}: ConVIRT is a self-supervised, one-to-one VLM trained on chest X-rays and associated radiology reports. We use the implementation of ConVIRT provided in ViLMedic \cite{vilmedic2022}, which is trained on MIMIC-CXR. 
\item \textit{BioViL} \cite{biovil}: BioViL is a one-to-one VLM trained on chest X-rays and associated radiology reports. BioViL is trained using two self-supervised loss functions: (1) a standard image-text contrastive loss as well as (2) a masked language modeling loss to improve the quality of the text encoder. BioViL is trained on MIMIC-CXR. 
\item \textit{GLoRIA} \cite{huang2021gloria}: GLoRIA is a VLM trained on chest X-rays and radiology reports. GLoRIA leverages two self-supervised loss functions: (1) a standard image-text contrastive loss for global alignment between images and text as well as (2) a local contrastive loss for finer-grained alignment between image patches and words. GLoRIA is not a one-to-one model. We evaluate three variants of GLoRIA: (1) GLoRIA-Global Only, which represents an image with a single image-level embedding, (2) GLoRIA-Local Only, which uses all image patch embeddings, and (3) GLoRIA-Global+Local, which uses both the image-level embedding as well as all patch embeddings. We use the implementation of GLoRIA provided in ViLMedic \cite{vilmedic2022}, which is trained on MIMIC-CXR. 
\end{itemize}

\subsection{Extended Details on Evaluating Region-Attribute Mappings}
\label{appenddixsec:evaluatingregatt}
\noindent
In Section \ref{sec:mappingaccresults}, we evaluated the quality of the region-attribute pairs generated using the mapping model. Here, we provide additional implementation details related to these evaluations.

\smallskip
\smallskip
\noindent
As described in Section \ref{subsec:train}, we use our trained mapping model (Stage 1 of \name{}) to generate region-attribute pairs. Our goal here is to quantitatively measure the number of generated region-attribute pairs that capture correct associations. However, measuring region-attribute quality requires ground-truth pairings, which are not always provided in real-world datasets. As a result, we generate ground-truth region-attribute pairings for each of our four pretraining datasets as follows:
\begin{itemize}
\item \textit{DocMNIST}: Since \textsc{DocMNIST} is a synthetically-generated dataset, we have access to the ground-truth attributes associated with each region.
\item \textit{DeepFashion}: DeepFashion provides human parsing labels for a subset of the images, where various features of each image (\eg glasses, hat, etc.) are labeled with pixel-level annotations in the image. We use our computed regions (four per image) and the human parsing labels to assign attributes to their ground-truth regions. 
\item \textit{MIMIC-CXR}: MIMIC-CXR does not provide any region-level annotations. As a result, we estimate region-attribute quality by focusing only on two entities - cardiomegaly and pacemaker; these entities were selected due to their consistent association with a single region: the heart. We evaluate mapping accuracy using only this subset of two ground-truth region-attribute pairs. In the future, we aim to conduct user studies to better evaluate the quality of region-attribute mappings on datasets like MIMIC-CXR that lack region-level annotations.
\item \textit{COCO}: On the COCO dataset, we train our mapping model using candidate regions generated by a RPN; however, we do not have have access to ground-truth attribute pairings for these generated regions. As a result, we estimate region-attribute mapping quality by using the annotated, ground-truth regions in the COCO dataset. We evaluate the number of ground-truth regions that can be accurately mapped to their corresponding attributes.
\end{itemize}

\smallskip
\smallskip
\noindent
For each image-text sample, we are given a set of generated region-attribute mappings as well as a set of ground-truth region-attribute pairs. A region-attribute pair is correct if it occurs in both sets. We compute \textit{mapping precision} and \textit{mapping recall} with the following formulas:
$$\text{mapping precision} = \frac{\text{number of correct region-attribute mappings}}{\text{number of generated region-attribute mappings}}$$
$$\text{mapping recall} = \frac{\text{number of correct region-attribute mappings}}{\text{number of ground-truth region-attribute pairs}}$$

\smallskip
\smallskip
\noindent
As our final quality metric, we report the \textit{mapping F1} score in Table \ref{tab:mapping_acc}, which is computed using the standard formula for F1 scores:
$$\text{mapping F1} = \frac{2*\text{mapping precision}*\text{mapping recall}}{\text{mapping precision} + \text{mapping recall}}$$

\smallskip
\smallskip
\noindent
We note here that there exists a tradeoff between mapping precision and mapping recall that is related to the choice of $\epsilon$ in Section \ref{subsec:train}. If $\epsilon$ is high, then \name{} will generate a large number of region-attribute pairs, resulting in a low mapping precision and a high mapping recall. If $\epsilon$ is low, then \name{} will generate a small number of region-attribute pairs, resulting in a high mapping precision and a low mapping recall. Empirically, as stated in Appendix Section \ref{appendixsec:stage2details}, we select the value of $\epsilon$ by computing mapping F1 scores on a validation set (if sufficient ground-truth region-attribute pairings are available). We note here that when ground-truth region-attribute pairings are limited (such as in the MIMIC-CXR dataset), we set $\epsilon$ to a default value of 0.1.

\end{document}